\pdfoutput=1

\documentclass[11pt]{article}

\usepackage{acl}

\usepackage{times}
\usepackage{latexsym}

\usepackage[T1]{fontenc}

\usepackage[utf8]{inputenc}

\usepackage{microtype}

\usepackage{inconsolata}

\usepackage{graphicx}
\usepackage{amsmath}
\usepackage{tikz}
\usepackage{pgfplots}
\usepackage{subfigure}
\usepackage{subcaption}
\usepackage{amssymb}
\usepackage{booktabs}

\newcommand{\methodname}{\textit{EMBER} }
\newcommand{\methodnamelower}{ember }
\newcommand{\methodnamepunctuation}{\textit{EMBER}}
\newcommand{\methodnamefull}{\textit{Embedded Named Entity Recognition} }

\title{Embedded Named Entity Recognition using Probing Classifiers}

\author{Nicholas Popovi\v{c}\textsuperscript{1} \and Michael Färber\textsuperscript{2}\\
  \textsuperscript{1}Karlsruhe Institute of Technology, Germany \textsuperscript{2}TU Dresden \& ScaDS.AI, Germany \\
  \texttt{popovic@kit.edu, michael.faerber@tu-dresden.de} \\}

\begin{document}
\maketitle
\begin{abstract}

Streaming text generation has become a common way of increasing the responsiveness of language model powered applications, such as chat assistants.
At the same time, extracting semantic information from generated text is a useful tool for applications such as automated fact checking or retrieval augmented generation.
Currently, this requires either separate models during inference, which increases computational cost, or destructive fine-tuning of the language model.
Instead, we propose an approach called \methodname which enables streaming named entity recognition in decoder-only language models without fine-tuning them and while incurring minimal additional computational cost at inference time.
Specifically, our experiments show that \methodname maintains high token generation rates, with only a negligible decrease in speed of around 1\% compared to a 43.64\% slowdown measured for a baseline.
We make our code and data available online\footnote{\url{https://github.com/nicpopovic/EMBER}}, including a toolkit\footnote{\url{https://github.com/nicpopovic/STOKE}} for training, testing, and deploying efficient token classification models optimized for streaming text generation.

\end{abstract}

\section{Introduction}

Combining pre-trained language models (LMs) and external information at inference time is a widely used approach, for example as a means of improving the factual accuracy of generated texts in knowledge-intensive tasks \cite{NEURIPS2020_6b493230, pmlr-v119-guu20a, gaoRetrievalAugmentedGenerationLarge2024}.
\begin{figure}[t]
    \centering
    \includegraphics[width=0.46\textwidth]{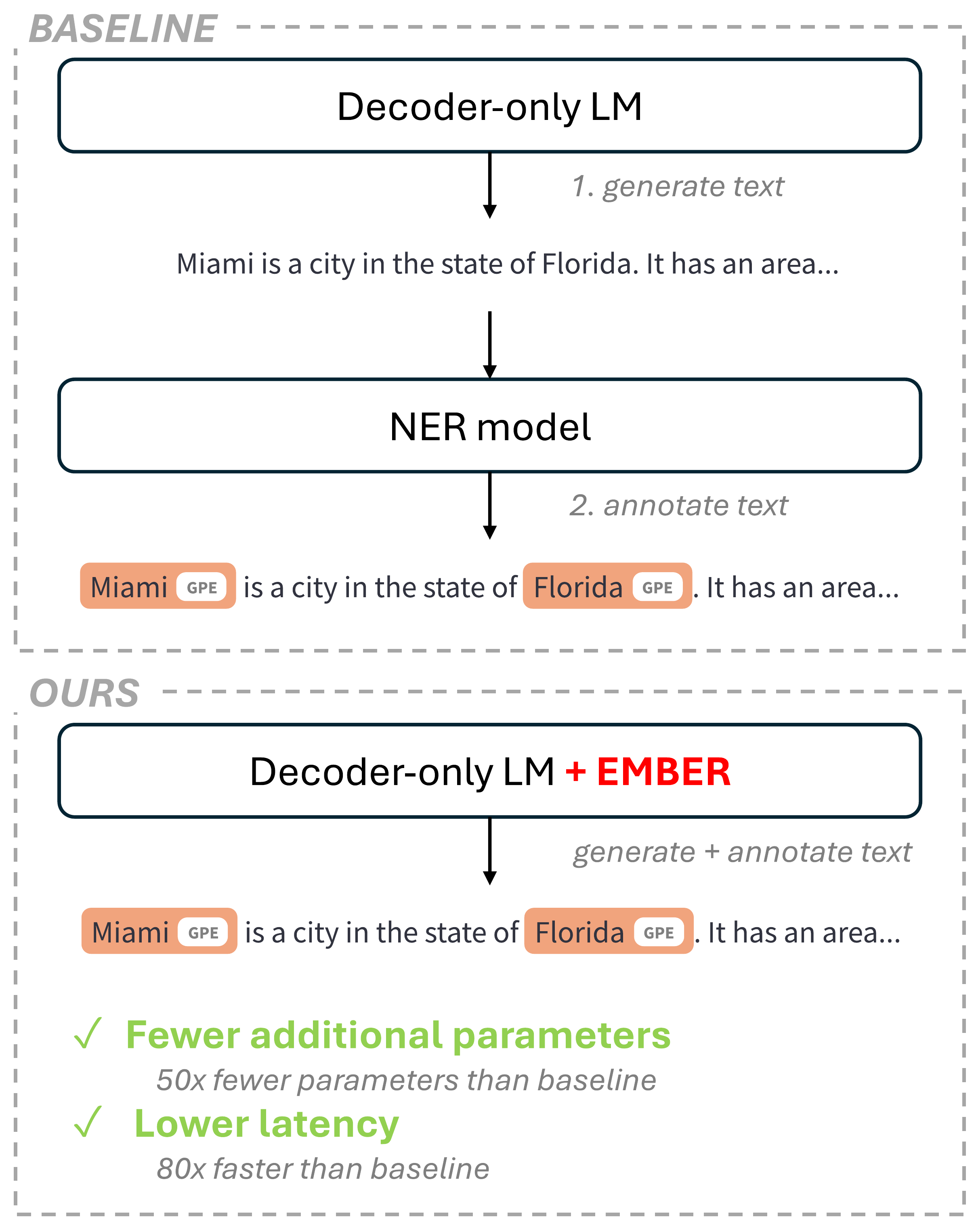}
    \caption{\methodname enables simultaneous text generation and entity annotation by using a language model's internal representations as the feature space for classification. Compared to using state-of-the-art NER models, this results in a substantially more efficient pipeline allowing for streaming named entity recognition. Parameter and latency comparisons stated in this figure are based on the experiments conducted using GPT-2$_{\text{XL}}$, presented in section \ref{sec:simgenNER}.}
    \label{fig:teaser}
\end{figure}
To effectively gather relevant information, there are generally two main strategies for extracting semantic data from the current context, each with its own drawbacks. The first strategy involves integrating the extraction process into text generation. This method, as seen in work by \citet{schickToolformerLanguageModels2023a} and \citet{Zhang2023GraphToolFormerTE}, requires generating queries during the inference phase. Although this approach is direct, it has the downside of altering the LM through fine-tuning, which can lead to issues such as catastrophic forgetting (\citet{goodfellowEmpiricalInvestigationCatastrophic2015}). The second strategy employs an external system for information extraction (IE). While studies such as those by \citet{shiREPLUGRetrievalAugmentedBlackBox2023}, \citet{ram-etal-2023-context}, and \citet{dhuliawalaChainofVerificationReducesHallucination2023} show promising results, the required computational overhead is a major issue hindering adoption \cite{chen2023purr, zhang2023sirens}.
For many applications (chat assistants, etc.), the delivery of generated text on a token-by-token basis as soon as they are available, known as streaming text generation, has become a common way of increasing responsiveness.
An optimized solution for this setting is currently missing.\\

Meanwhile, research into the mechanistic interpretability of LMs has shown that substantial semantic information can be recovered from individual internal representations.
A common diagnostic tool are simple\footnote{Typically small in the amount of trainable parameters and less complex in terms of architecture relative to the LM.} classifiers, called probing classifiers \cite{belinkov-glass-2019-analysis}, trained to perform specific tasks using a subset of the internal representations of a (frozen) LM as their feature space.
While their validity as a means for understanding how and where information is stored in LMs is debated \cite{cao-etal-2021-low, belinkov-2022-probing}, probing classifiers have been shown able to map internal representations to syntactic and semantic information \cite{raganato-tiedemann-2018-analysis, clark-etal-2019-bert, marecek-rosa-2019-balustrades, htutAttentionHeadsBERT2019, pimentel-etal-2020-information, schouten-etal-2022-probing}.
While the majority of this research has been conducted using encoder LMs, studies have shown that similar information is recoverable from specific internal states of decoder-only LMs \cite{mengLocatingEditingFactual2022c, gevaDissectingRecallFactual2023a, hernandezLinearityRelationDecoding2023a, ghandehariounPatchscopesUnifyingFramework2024}.
We therefore explore whether, rather than as a diagnostic tool, probing classifiers can be used for non-destructive, light-weight, and continuous IE in decoder-only LMs at inference time.\\
\begin{figure}[t]
    \centering
    \scalebox{0.92}{
    \includegraphics{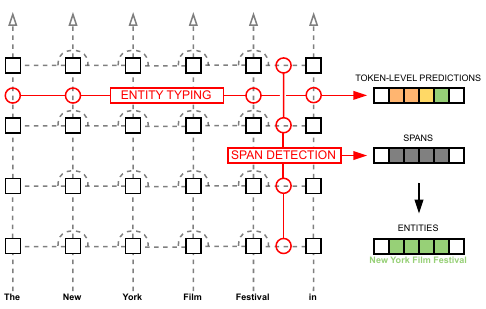}
    }
    \caption{Illustration of the proposed approach for named entity recognition using probing classifiers.
    Black squares symbolize individual transformer layers at individual timesteps, while dotted lines symbolize information flow throughout the transformer. Probing classifiers are shown in red, with circles symbolizing where representations are accessed. One classifier performs token-level entity typing using hidden states at a single layer, while second classifier detects spans based on attention weights. Both predictions are aggregated into span-level entity predictions.}
    \label{fig:main_figure}
\end{figure}
In this work, we develop an approach we call \methodnamefull (\methodnamepunctuation) for performing named entity recognition (NER), a central IE subtask consisting of mention detection and entity typing, using only a LM's internal representations as feature space without further finetuning thereof.
As illustrated in figure \ref{fig:main_figure}, the process involves two probing classifiers: 
The first performs tokenwise type classification based on the hidden state at a single transformer sublayer,
while the second detects spans based on the LMs attention weights between two tokens.
Finally, the outputs of both are fused into span-level entity predictions.
We conduct a series of experiments using multiple LMs, NER datasets, and task settings to evaluate the performance of \methodname and the factors which influence it.
In short, we find that while outperformed by finetuned state-of-the-art approaches in terms of raw benchmark scores ($\sim$$80-85\%$ F1 vs. $>$$90\%$ F1), our approach outperforms few-shot in-context learning approaches ($\sim$$50\%$ F1) (section \ref{sec:supervised}) and is significantly more efficient in the streaming text generation setting (approx. $80\times$ faster than baseline (section \ref{sec:simgenNER})).

In conclusion, we make the following contributions:
We propose \methodnamepunctuation, the first non-destructive NER approach optimized specifically for use with decoder-only LMs during streaming text generation.
We show that our approach can achieve F1 scores of 80-85\% while requiring minimal additional computational overhead.
We provide insight into which architecture parameters of decoder-only LMs determine how well our approach will work.
Lastly, we showcase efficient simultaneous text generation and NER, a novel usecase our approach is optimized for and provide a toolkit for training, testing, and deploying models.

\section{Related Work}

\subsection{NER using Pretrained Language Models}
As a long standing NLP task, researchers have tackled named entity recognition (NER) using a wide variety of approaches \cite{liSurveyDeepLearning2022}, with most state-of-the-art approaches relying on fine-tuning pretrained encoder language models \cite{Luo_Xiao_Zhao_2020, fu-etal-2021-spanner, wang-etal-2021-automated, ye-etal-2022-packed}. \citet{wang-etal-2021-automated} use an ensemble-approach to show that non-fine-tuned embeddings can also be feasible. Parameter efficient fine-tuning (PEFT) aims to significantly reduce the amount of parameters trained by using low-rank adaptations \cite{hu2021lora}, prompt-tuning \cite{shen-etal-2023-promptner} or adapters \cite{nie-etal-2024-know-adapter}. While our proposed approach is conceptually similar to adapters, PEFT approaches aim to emulate destructive finetuning, while our approach is non-destructive. With respect to generative language models, existing approaches typically frame the task as a sequence generation task, where the model outputs a sequence of entities for a given text either through fine-tuning \cite{ijcai2021p542, yan-etal-2021-unified-generative, lu-etal-2022-unified, josifoski-etal-2022-genie}, or in an in-context learning setting \cite{epure-hennequin-2022-probing, wang2023gpt, chen-etal-2023-learning, ashok2023promptner, guo2023retrievalaugmented, li-etal-2023-codeie}.

\subsection{Language Models and Probing Classifiers}
Considerable research using probing classifiers to predict linguistic properties based on a LM's internal representations\footnote{Note that the term probing is also used for analyses conducted in an in-context learning setting (see for example \citet{epure-hennequin-2022-probing}), a parameter-free technique which differs from the use probing classifiers.}, including those related to entities, has been conducted \cite{ettinger-etal-2016-probing, shi-etal-2016-string, adi2017finegrained, tenney2018what, belinkov-glass-2019-analysis} with studies primarily being applied to encoder or encoder-decoder LMs. 
Probing classifiers are typically used as a diagnostic tool for understanding information storage or flow in LMs. 
As such, most recent studies opt for less complex, often linear probes in order to prevent the representational capabilities of the probe from falsifying results \cite{cao-etal-2021-low, belinkov-2022-probing}.
Recently, decoder-only LMs appear to have become more popular than other architectures for many tasks, likely due to increased availability of larger pretrained models \cite{brownLanguageModelsAre, workshopBLOOM176BParameterOpenAccess2022, zhangOPTOpenPretrained2022, chowdheryPaLMScalingLanguage2022, touvronLLaMAOpenEfficient2023} and the flexibility offered by the generative framing of many tasks, for example as in-context learning.
Interpretability research focusing on decoder-only LMs has shown that similar to encoder LMs, semantic information is recoverable from specific internal states of decoder-only LMs \cite{mengLocatingEditingFactual2022c, gevaDissectingRecallFactual2023a, hernandezLinearityRelationDecoding2023a, wang-etal-2023-label, ghandehariounPatchscopesUnifyingFramework2024}. 

\subsection{Streaming Token Classification}
To the best of our knowledge, our approach is the first dedicated, non-destructive solution to streaming token classification for generative language models.
Existing token classification pipelines are not designed to process information incrementally, but instead expect a completed text as input.
This means that classification must either be performed after the text generation is completed, complicating streaming output delivery, or that the generation will be slowed down substantially, as the full text must be re-processed at every increment.

\section{Task Description}
NER consists of two subtasks, namely mention detection and entity typing.
Given a text as a sequence of tokens $t_{1},...t_{N}$, and a set of entity types $E$, mention detection involves locating all spans $t_{i},...t_{j}$, where $1 < i,j < N$, corresponding to mentions of entities within the text. 
Entity typing is the task of assigning the correct entity type $e \in E$ to each mention.
For Transformer-based approaches, NER is typically framed as a token classification task, where each token $t_{i}$ is assigned a label $y_{i}$ based on whether it is the first token of a mention (B), inside a mention (I) or outside of a mention (O).

\section{\methodnamepunctuation}
\label{sec:approach-overview}

In this section we introduce our approach for building a NER system based on probing internal representations of pretrained, decoder-only LMs.
Given a model $M$ with $L$ layers, a hidden state $h_{i}^{l}$ is the output at a single transformer sublayer, where $l\in [1,...,L]$ is the index of the sublayer and $i$ is the index of the input token.
The attention weights\footnote{In contrast to the hidden state probes which are restricted to a single sublayer at a time, attention probes use the weights for all attention heads across all layers.} between two tokens are denoted as $A_{j, i}$, where $j \geq i$ due to autoregressivity.
Our approach entails two key steps, namely tokenwise entity type classification based on $h_{i}^{l}$ (\ref{sec:approach-typing}) and span detection based on $A_{j, i}$(\ref{sec:approach-spandetection}), the results of which are then combined to form a complete NER pipeline using a mechanism we call label propagation (\ref{sec:approach-pipeline}).

\subsection{Tokenwise Classification}
\label{sec:approach-typing}
Prior work has shown that individual hidden states contain sufficent information to recover semantic information about entities, suggesting that these may represent a suitable feature space for our goal of entity typing.
We therefore perform tokenwise classification by learning $f_{type}$ such that:
\begin{equation}
    f_{type}(h_{i}^{l}) = \hat{y}_{i},
\end{equation}
where $\hat{y}_{i}$ is a prediction in IOB2 format.\\

\noindent
\textbf{However,} the autoregressive nature of decoder-only LMs results in two inherent issues we will address in the following.
(1) Firstly, entities which have their type modified by context following the mention cannot be correctly classified (For example, in the phrase \textit{``Harry Potter is a book title.''}, \textit{``Harry Potter''} may be classified as a person given only the initial two words, while the remaining context makes the assignment of a type such as ``work of art'' more suitable. See \ref{sec:appendix-examples} for an illustration of this example.).
While, this issue is an inherent limitation of \methodnamepunctuation , our experiments show that its impact on general NER performance is limited.
(2) The second issue arises for entities spanning multiple tokens.
Consider the composite phrase \textit{``New York Film Festival''} (see also figure \ref{fig:main_figure} and appendix \ref{sec:appendix-examples}), which, given the annotation schema for Ontonotes5 \cite{hovy-etal-2006-ontonotes}, should be assigned the entity type \textit{``EVENT''}.
Given only the partial phrase \textit{``New York''}, however, the most appropriate entity type to assign is \textit{``GPE''}.
We therefore expect that a token-level classifier outlined above will not predict all tokens in this phrase as belonging to the class \textit{``EVENT''}.
More generally, classifying on a per-token basis does not guarantee that the same class is assigned to all tokens within a mention span.
In the following section, we therefore provide a method for detecting entity spans, using which we can then aggregate tokenwise predictions to span-level predictions.

\begin{figure}[t]\hfill
    \centering
    \begin{subfigure}
        \centering
        \includegraphics[width=0.9\linewidth]{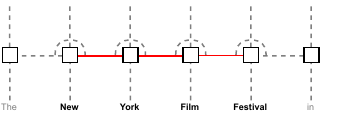}
        \vspace*{-5mm}
        \caption*{(a) neighbour classification}
        \vspace*{5mm}
    \end{subfigure}
    \hfill
    \begin{subfigure}
        \centering
        \includegraphics[width=0.9\linewidth]{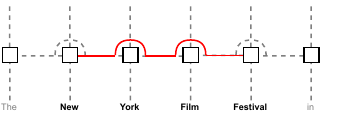}
        \vspace*{-5mm}
        \caption*{(b) span classification $j=k$}
        \vspace*{5mm}
    \end{subfigure}
    \hfill
    \begin{subfigure}
        \centering
        \includegraphics[width=0.9\linewidth]{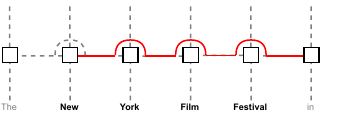}
        \vspace*{-5mm}
        \caption*{(c) span classification $j=k-1$}
    \end{subfigure}
    \caption{Illustration of the different span detection methods. Red colors indicate which attention weights to classify as positive for the example span ``New York Film Festival''. Attention weights are only shown for a single layer, but are generally used at all layers.}
    \label{fig:span_detection}
\end{figure}

\subsection{Span Detection}
\label{sec:approach-spandetection}
Since attention is the mechanism by which decoder-only LMs incorporate information from preceeding tokens, we hypothesize that $A_{j, i}$ contains different information based on whether or not $M$ represents $t_i$ and $t_j$ as tokens within the same span.
Below, we propose two different approaches for identifying spans based on $A_{j, i}$:\\

\noindent
\textbf{Neighbour Classification.} In neighbour classification, illustrated in figure \ref{fig:span_detection} (a), we train a classifier to predict whether two adjacent tokens belong to the same mention ($a_{i,j} = 1$ if so, $a_{i,j} = 0$ otherwise), based on the attention weights $A_{j, i}$, where $j = i+1$.
\begin{equation}
    f_{adj}(A_{j, i}) = \hat{a}_{i,j},
\end{equation}
\noindent
\textbf{Span Classification.} For span classification, illustrated in figures \ref{fig:span_detection} (b)+(c), we train a classifier to predict, based on $A_{k, i}$, whether $i$ is the first and $j$ is the last token of the same mention ($s_{i,j} = 1$ if so, $s_{i,j} = 0$ otherwise):
\begin{equation}
    f_{span}(A_{k, i}) = \hat{s}_{i,j},
\end{equation}
where either  $j=k$ (figure \ref{fig:span_detection}.b) or $j=k-1$ (figure \ref{fig:span_detection}.c), the reason behind the latter being autoregressivity:
Without seeing the next token, it is not always possible to confidently predict whether the current token is the last of a given span (\textit{``New York''} could be part of a span such as \textit{``New York Film Festival''}).

\subsection{Label Propagation}
\label{sec:approach-pipeline}
Having generated predictions for the types of individual tokens and for which tokens make up a span, the final step is to combine the two sets of information into NER predictions.
Rather than applying a voting or pooling mechanism to decide which entity type prediction is the correct one for a span containing multiple tokens, we choose the type predicted for the last token of a span, as for this index $M$ has access to the largest amount of context\footnote{Prior work also reports information aggregation to later tokens \cite{gevaDissectingRecallFactual2023a, wang-etal-2023-label}.}.
We refer to this as label propagation.
Our experiments (appendix \ref{sec:appendix-entitytyping}) show that high F1 scores can be achieved by using solely the type assigned to the last token.
Below, we propose three different approaches for label propagation:
\subsubsection*{Adjacency-based Propagation}
In adjacency-based propagation, we iterate over all tokenwise predictions $\hat{Y}$ in descending order (referring to the sequence index).
If $\hat{y}_{i} \neq ``O''$ and $\hat{a}_{i-1, i} > 0$, we assign $\hat{y}_{i-1} = \hat{y}_{i}$.

\subsubsection*{Spanwise Typing}
For spanwise typing, we select all spans for which $\hat{s}_{i, j} > 0$ (for overlapping values we chose the span with the highest $\hat{s}_{i, j}$).
For the resulting spans, we select as entity type $\hat{y}_{j}$. Where $\hat{y}_{j} = ``O''$, we chose the second most likely type in order to guarantee that an entity type is assigned.

\subsubsection*{Spanwise Propagation}
In span-based propagation, we again iterate over all tokenwise predictions $\hat{Y}$ in descending order.
If $\hat{y}_{i} \neq ``O''$ and $\{j \in \mathbb{R}: \hat{s}_{i, j} > 0\}$, we select the most likely span $j_{\text{max}} = \arg\max_{j \in \mathbb{R}} \hat{s}_{i, j}$ and assign $\hat{y}_{k} = \hat{y}_{i}$ for all $k \in [j_{\text{max}}, ..., i]$.

\section{Experiments: Non-Streaming NER}

Before examining the novel task setting of streaming named entity recognition (NER), we conduct experiments to determine how well \methodname performs in a variety of typical, non-streaming NER settings.
We begin by evaluating which label propagation strategies work best (\ref{sec:ablation}).
After identifying the best configuration, we evaluate its performance in the supervised learning setting (\ref{sec:supervised}).
Next, we analyse the measured results with respect to the effects of model scale and architecture parameters (\ref{sec:scaling}).
Finally, we evaluate our approach in heavily data-constrained settings (\ref{sec:fewshot}) and show the efficient extraction of named entities during streaming text generation (\ref{sec:simgenNER}).\\

\noindent
\textbf{Models and Data.}
All experiments are performed using the datasets CoNLL2003 \cite{tjong-kim-sang-de-meulder-2003-introduction} and Ontonotes5 \cite{hovy-etal-2006-ontonotes} and 7 LMs from the model families GPT-2 \cite{radford2019language}, GPT-J \cite{gpt-j}, and Pythia \cite{biderman2023pythia}.
Further details are provided in the individual sections and appendix \ref{sec:appendix-implementation-details}.\\

\subsection{Label Propagation Strategies}
\label{sec:ablation}

We train probing classifiers as introduced in \ref{sec:approach-typing} and \ref{sec:approach-spandetection} in the supervised setting and compare the results of the different label propagation approaches introduced in \ref{sec:approach-pipeline}. Results shown in this section are the top results obtained on the validation splits of the datasets. 
We show only the results for GPT-2$_\text{XL}$.
Results for other models exhibit the same trends and are included in appendix \ref{sec:appendix-spandetection}.\\

\noindent
\textbf{Results.} 
\begin{table}[t]
\centering
\scalebox{0.81}{
\begin{tabular}{lcccc}
\textbf{Approaches} & MD & P & R & F1 \\ \hline

\hline
\multicolumn{5}{c}{\textsc{conll2003}} \\
\hline
(Tokenwise typing) & H & 64.55\% & 79.10\% & 71.09\% \\
Adj. propagation & H & 84.40\% & 90.47\% & 87.33\% \\
Spanwise typing & A & 89.32\% & \textbf{91.75\%} & \textbf{90.52\%} \\
Span propagation & H+A & \textbf{94.08\%} & 87.13\% & 90.47\% \\

\hline
\multicolumn{5}{c}{\textsc{ontonotes5}} \\
\hline
(Tokenwise typing) & H & 58.56\% & 71.55\% & 64.41\% \\
Adj. propagation & H & 68.25\% & 76.11\% & 71.96\% \\
Spanwise typing & A & 76.79\% & \textbf{76.26}\% & 76.52\% \\
Span propagation & H+A &\textbf{87.26\%} & 72.77\% & \textbf{79.36\%} \\

\end{tabular}
}
\caption{NER scores for GPT-2$_\text{XL}$ using hidden states and attention weights in different ways. The column ``MD'' indicates the feature space used for mention detection in the approach, where ``H'' stands for hidden state and "A" stands for attention. All scores are micro F1 scores measured on the validation sets of CoNLL2003 and Ontonotes5.}
\label{tab:results-ablation}
\end{table}

In table \ref{tab:results-ablation} we show precision, recall, and F1 scores for the different NER approaches.
For reference, we include the results for tokenwise classification where we measure F1 scores of $71.09\%$ and $64.41\%$, further highlighting the need for label propagation.
As for the label propagation variants outlined in \ref{sec:approach-pipeline}, we find that span propagation tends to lead to the highest F1 scores on Ontonotes5 ($79.36\%$) and the second highest (by a close margin) for CoNLL2003 90.47\%.
Span propagation exhibits significantly higher precision than recall, since it requires both classifiers to detect an entity for mention detection (see ``MD'' in table \ref{tab:results-ablation} for a comparison of the active mention detection mechanisms).\\

\noindent
\textbf{Conclusion.} Based on the above results, we select spanwise propagation using the span detected based on $j=k-1$ for all following experiments. 

\subsection{Supervised Learning}
\label{sec:supervised}
We evaluate \methodname on the test sets of the two benchmarks.
For lack of directly comparable approaches, we include two different types of baselines representing the use of external extraction mechanisms at inference time:
In order to provide an upper bound for F1 scores that can be achieved on each dataset we select state-of-the-art approaches based on finetuning encoder language model architectures \cite{wang-etal-2021-automated, ye-etal-2022-packed}.
Secondly, we include results for GPT-2$_\text{XL}$ and GPT-J$_\text{6B}$ in an in-context learning 5-shot setting \cite{chen-etal-2023-learning}.
While the heavy data-constraints of the 5-shot setting result in an unfair comparison on the surface, in-context learning is the prevalent method for using decoder-only language models without finetuning and necessarily limits the amount of data which can be used due to context size limitations.
We show the results for the largest model of each model family.
Further results and details are given in appendices \ref{sec:appendix-implementation-details} and \ref{sec:appendix-full-results}. 
In appendix \ref{sec:appendix-further-datasets} include results for the datasets WNUT2017 \cite{derczynski-etal-2017-results} and BC5CDR \cite{liBioCreativeCDRTask2016a} and in appendix \ref{sec:appendix-llama} we include results for newer LMs (Llama-3.2 \cite{dubey2024llama3herdmodels}).\\

\begin{table}[t]
\centering
\scalebox{0.9}{
\begin{tabular}{l|c|cc}
\textbf{Model} & \textbf{param$_\text{add.}$} & \multicolumn{1}{c}{\textbf{CoNLL2003}} & \multicolumn{1}{c}{\textbf{Ontonotes5}} \\
\hline
\multicolumn{3}{c}{\textsc{sota (finetuned)}} \\
\hline
ACE & > 500M* & 94.6\% &-\\

PL-Marker & 355M &-& 91.9\% \\
\hline
\multicolumn{3}{c}{\textsc{5-shot icl \cite{chen-etal-2023-learning}}} \\
\hline
GPT-2$_\text{XL}$ & 0 & 39.55\% &-\\
GPT-J$_\text{6B}$ & 0 & 50.10\% &-\\
\hline
\multicolumn{3}{c}{\textsc{\methodnamelower (ours)}} \\
\hline
GPT-2$_\text{XL}$ & 11.5M & 85.14\% & 79.26\% \\
GPT-J$_\text{6B}$ & 18.6M & 83.68\% & 76.70\% \\
Pythia$_\text{6.9b}$ & 21M & 83.90\% & 78.85\% \\

\end{tabular}
}
\caption{NER F1 scores for CoNLL2003 and Ontonotes5 in the supervised learning setting. Results for PL-Marker as reported by \citet{ye-etal-2022-packed}, results for ACE as reported by \citet{wang-etal-2021-automated}. param$_\text{add.}$ indicates the number of parameters dedicated only to NER that are required for each approach. * For ACE, since it is an ensemble approach, the number of parameters can vary, so we give an estimate based on the configuration reported by the authors.}
\label{tab:results-md}
\end{table}

\noindent
\textbf{Results.} As shown in table \ref{tab:results-md}, we measure F1 scores in the range of $83.68-85.14\%$ for CoNLL2003 and $76.70-79.26\%$ for Ontonotes5.
In both cases, these results are below the state-of-the-art results for encoder style models ($94.6\%$ for CoNLL2003, $91.7\%$ for Ontonotes5).
We observe that mention detection appears to be the main bottleneck for \methodnamepunctuation, with detailed results in appendix \ref{sec:appendix-et-v-md}.
Compared to the in-context learning baseline, however, the scores are significantly higher ($+45.59\%$ for GPT-2$_\text{XL}$ and $+33.10\%$ for GPT-J$_\text{6B}$).
When comparing the results of the different LMs to their model sizes it becomes apparent that GPT-2$_\text{XL}$ exhibits higher F1 scores than both GPT-J$_\text{6B}$ and Pythia$_\text{6.9b}$, even though it has considerably fewer parameters.
We expand on this observation in the following section.\\

\noindent
\textbf{Conclusion.} Our experiments show that in a non-streaming NER setting, existing state-of-the-art approaches outperform our approach w.r.t. to annotation quality. This is unsurprising, since \methodname does not involve finetuning of the LMs representations. The results do, however, also show that our approach is capable of performing NER with F1 scores of up to approx. 85\% (outperforming, for example, in-context learning).

\subsection{Effects of Architecture \& Scale}
\label{sec:scaling}
On the surface, the observation that GPT-2$_\text{XL}$ performs better than models with approx. 4 times the amount of parameters runs contrary to the intuition that models with more parameters result in better representational capabilities.
When considering the differences in architectures of the three LMs (details of which can be found in appendix \ref{sec:appendix-modelspecs} table \ref{tab:model-specs}), however, we find that GPT-2$_\text{XL}$ has the highest number of attention heads ($1200$ vs. $1024$/$448$).
Since \methodname uses the attention weights as feature space for span detection, the fact that the feature space has the highest dimensionality for GPT-2$_\text{XL}$ provides a possible explanation.
We, therefore, investigate the effects of hidden state and attention weight dimensionality on F1 scores using 7 LMs, ranging from 125m to 6.9b parameters in size.\\

\begin{figure}[t]
\centering
\scalebox{0.83}{
    \begin{tikzpicture}
    \begin{axis}[legend pos=south east, ylabel=F1\%, xlabel=\# hidden units, only marks, xshift=-5cm]
    \node[rotate=90, anchor=center, text=gray] at (axis cs:768,93.01078112579685) {GPT-2};
    \node[rotate=90, anchor=center, text=gray] at (axis cs:1024,93.86871452050222) {Py-410m};
    \node[rotate=90, anchor=center, text=gray] at (axis cs:4096,94.84772768486842) {\begin{minipage}{1.5cm}GPT-J 6b \\ Py-6.9b\end{minipage}};
    \node[rotate=90, anchor=center, text=gray] at (axis cs:2560,94.39415646508331) {Py-2.8b};
    \node[rotate=90, anchor=center, text=gray] at (axis cs:2048,94.33430322869856) {Py-1.4b};
    \node[rotate=90, anchor=center, text=gray] at (axis cs:1600,94.09077039248552) {GPT-2 XL};
    \addplot[mark=+, color=blue] table [x=x, y=y, col sep=comma] {plot_entity_typing_conll.csv};\addlegendentry{CoNLL2003}
    \addplot[mark=x, color=black] table [x=x, y=y, col sep=comma] {plot_entity_typing_onto.csv};\addlegendentry{Ontonotes5}
    \end{axis}
    \end{tikzpicture}
    }
\caption{Entity typing F1 scores (validation set) for models with respect to hidden state dimension.}
    \label{fig:scale-entitytyping}
\end{figure}
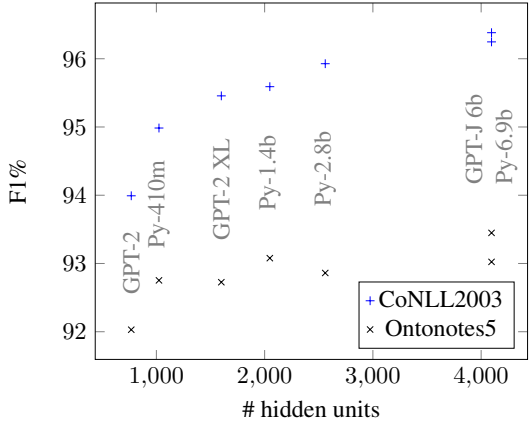

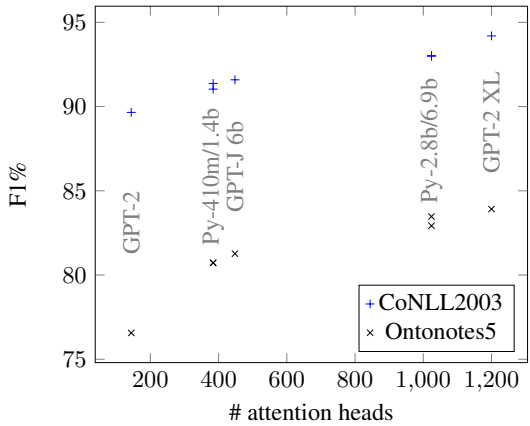
\begin{figure}[t]
\centering
\scalebox{0.83}{
    \begin{tikzpicture}
    \begin{axis}[legend pos=south east, ylabel=F1\%, xlabel=\# attention heads, only marks, xshift=-5cm]
    \node[rotate=90, anchor=center, text=gray] at (axis cs:144,83.1088500280193) {GPT-2};
    \node[rotate=90, anchor=center, text=gray] at (axis cs:384,85.87541948149166) {Py-410m/1.4b};
    \node[rotate=90, anchor=center, text=gray] at (axis cs:1024,87.97501740240622) {Py-2.8b/6.9b};
    \node[rotate=90, anchor=center, text=gray] at (axis cs:448,86.42880580756193) {GPT-J 6b};
    \node[rotate=90, anchor=center, text=gray] at (axis cs:1200,89.0560236945653) {GPT-2 XL};
    \addplot[mark=+, color=blue] table [x=x, y=y, col sep=comma] {plot_mention_detection_conll.csv};\addlegendentry{CoNLL2003}
    \addplot[mark=x, color=black] table [x=x, y=y, col sep=comma] {plot_mention_detection_onto.csv};\addlegendentry{Ontonotes5}
    \end{axis}
    \end{tikzpicture}
    }
\caption{Mention detection F1 scores (validation set) for models with respect to the total number of attention heads.}
    \label{fig:scale-spandetection}
\end{figure}

\noindent
\textbf{Results.} In figure \ref{fig:scale-entitytyping} we show a plot of the entity typing F1 scores measured for the different LMs compared to the dimensionality of the hidden states, which are the feature space for the corresponding probing classifier. 
We observe a clear, positive correlation between the two.
In figure \ref{fig:scale-spandetection} we show a plot of the mention detection F1 scores measured for the different LMs compared to the total number of attention weights.
Again, we observe a clear, positive correlation between the two.
When considering the absolute differences in F1 scores between the best and worst performing LMs for each task (CoNLL2003: $\Delta_{\text{ET}} = 2.39\%$, $\Delta_{\text{MD}} = 4.55\%$; Ontonotes5: $\Delta_{\text{ET}} = 1.42\%$, $\Delta_{\text{MD}} = 7.36\%$), we find that the effect of the total number of attention weights on mention detection is higher than that of the hidden state dimension on entity typing.

Lastly, we compare the NER F1 scores for Pythia$_\text{410m/1.4b}$ and Pythia$_\text{2.8b/6.9b}$, which have an identical number of attention heads at substantially different total model sizes (see table \ref{tab:results-scaling}). 
We see that they achieve nearly identical results, providing further evidence supporting the hypothesis that attention head count has a greater impact on \methodname at this scale of LM.\\

\begin{table}[t]
\centering
\scalebox{0.88}{
\begin{tabular}{lccc}
\textbf{Model} & \textbf{$|A|$} & \textbf{CoNLL2003} & \textbf{Ontonotes5} \\
\hline
Pythia$_\text{410m}$ & $384$ & $81.67\%$ & $76.54\%$ \\
Pythia$_\text{1.4b}$ & $384$ & $82.27\%$ & $76.59\%$ \\
\hline
Pythia$_\text{2.8b}$ & $1024$ & $84.63\%$ & $79.09\%$ \\
Pythia$_\text{6.9b}$ & $1024$ & $83.90\%$ & $78.85\%$ \\

\end{tabular}
}
\caption{NER micro F1 scores for CoNLL2003 and Ontonotes5 for 4 Pythia models. $|A|$ denotes hidden state dimensionality.}
\label{tab:results-scaling}
\end{table}

\noindent
\textbf{Conclusion.} We find that for this scale of LMs, the number of attention heads is a greater indicator of the overall performance of \methodname than the hidden state dimensionality.

\subsection{Few-Shot Learning}
\label{sec:fewshot}
Having evaluated \methodname in a supervised learning setting, we now evaluate it in a low-data setting using CoNLL2003 \cite{tjong-kim-sang-de-meulder-2003-introduction}.

We primarily report the results for GPT-2$_\text{XL}$ and GPT-J$_\text{6B}$, as these are the models for which we have in-context learning comparisons.
Data for all other models has been collected and is included in appendix \ref{sec:appendix-fewshot}.
We note, that the evaluation used to obtain the in-context learning results \cite{chen-etal-2023-learning} is not precisely identical to the one used in our experiments and therefore view this baseline as a limited comparison, with only significant differences being indicative of trends.
Further details about the experiment setup are included in appendix \ref{sec:appendix-implementation-details}.\\

\begin{table}[t]
\centering
\scalebox{0.75}{
\begin{tabular}{l|ccccc}
\textbf{Model} & \textbf{1-shot} & \textbf{5-shot} & \textbf{10-shot} & \textbf{50-shot} & \textbf{100-shot} \\ \hline
\multicolumn{6}{c}{GPT-2$_\text{XL}$} \\
\hline
\textsc{icl*} & 33.69\% & 39.55\%  & - & - & - \\
\textsc{\methodnamelower} & 23.44\% & 39.59\% & 45.29\% & 54.81\% & 57.69\% \\

\hline
\multicolumn{6}{c}{GPT-J$_\text{6B}$} \\
\hline
\textsc{icl*} & 46.14\% & 50.10\% & - & - & - \\
\textsc{\methodnamelower} & 19.27\% & 34.88\% & 41.20\% & 52.34\% & 57.45\% \\

\end{tabular}
}
\caption{Few-Shot F1 scores for NER on CoNLL2003. All scores are micro F1 scores. *Results as reported by Chen et al. \cite{chen-etal-2023-learning}.}
\label{tab:results-fewshot}
\end{table}

\noindent
\textbf{Results.} In table \ref{tab:results-fewshot} we show the results.
For the 1-shot setting we find that in-context learning yields significantly better results for both GPT-2$_\text{XL}$ and GPT-J$_\text{6B}$.
In the 5-shot setting, in-context learning and \methodname perform equally well for GPT-2$_\text{XL}$, while for GPT-J$_\text{6B}$ in-context learning is again superior.
As in our previous experiments, GPT-2$_\text{XL}$ generally performs better than GPT-J$_\text{6B}$.
For $k$-shot settings $k \in [10,50,100]$ we find that the gap between GPT-2$_\text{XL}$ and GPT-J$_\text{6B}$ decreases for higher values of $k$.\\

\noindent
\textbf{Conclusion.} Overall
our experiments show the following:
(1) for extreme data constraints, where $k$ is low enough to fit all labelled data into an in-context learning prompt, in-context learning results in higher F1 scores than \methodnamepunctuation.
(2) In settings where $k$ is too high for in-context learning, yet too low for supervised learning, our approach presents a viable alternative.

\section{Experiments: Streaming NER}
\label{sec:simgenNER}
So far, all evaluation presented has involved NER on non-generated text.
\methodnamepunctuation, however, offers an efficient way of performing NER during text generation.
In the following we, therefore, set out to answer two questions:
What is the impact of using \methodname on inference speeds? 
Does it produce annotations of the same quality on generated text as it does on non-generated text?
\\

\noindent
\textbf{Dataset.} We begin by constructing an evaluation dataset by randomly sampling $50$ texts from the validation split of CoNLL2003 and using each as a prompt to generate text with GPT-2$_{\text{XL}}$.
We generate $100$ tokens for each prompt using greedy decoding with a repetition penalty \cite{keskarCTRLConditionalTransformer2019} of $1.2$ and manually annotate the resulting texts w.r.t. NER according to the CoNLL2003 annotation guideline.
In addition to the manually labelled evaluation dataset, we create synthetically labeled datasets for training and validation, by using a teacher model, XLM-RoBERTa$_\text{large}$ \cite{ruder-etal-2019-unsupervised}, for annotation.
We train another span detection probe on the synthetically generated data to compare its performance to a classifier trained on non-generated data.
Further details concerning the datasets are included in appendix \ref{sec:appendix-generated-dataset} and the toolkit used for their creation, as well as a model playground are available online.
As a baseline, we evaluate XLM-RoBERTa$_\text{large}$ on the dataset and compare performance on the generated and non-generated texts separately.
For the regular CoNLL2003 benchmark, the authors report an F1 score of $92.9\%$ for XLM-RoBERTa$_\text{large}$.\\

\begin{table}[t]
\centering
\scalebox{0.8}{
\begin{tabular}{lcc}
\textbf{Method} & \textbf{ms / token $\downarrow$} & \textbf{tokens / s $\uparrow$} \\ \hline
generation only & $28.12 \pm 0.15$ & $35.59 \pm 0.18$ \\
\hline
 + XLM-RoBERTa$_\text{large}$ & $49.96 \pm 0.07$ & $20.06 \pm 0.03$ \\
 $\Delta_\text{abs}$  & $+21.84 \text{ms}$ & $-15.53$ \\
$\Delta_\text{rel}$  & $+77.67 \%$ & $-43.64 \%$ \\
 \hline
 + \methodname & $28.39 \pm 0.13$ & $35.23 \pm 0.16$ \\
$\Delta_\text{abs}$  & $+0.27 \text{ms}$ & $-0.36$ \\
$\Delta_\text{rel}$  & $+0.96 \%$ & $-1.01 \%$ \\

\end{tabular}}
\caption{Impact of streaming NER during generation on inference speed for GPT-2$_\text{XL}$. The results show clearly how much more efficient \methodname is compared to the baseline approach, incurring a performance penalty on token generation rates of only 1\% (compared to more than 40\%).
}
\label{tab:time-impact}
\end{table}

\noindent
\textbf{Results - Efficiency.} In table \ref{tab:time-impact} we show the cost of performing NER after every generated token during text generation, which highlights the low computational overhead required for our approach.
We find that using \methodname slows down inference by only $0.27\text{ms}$ per token compared to $21.84\text{ms}$ for the baseline, reducing the amount of tokens generated per second by around 1\% compared to 43.64\% for the baseline.
When comparing the number of additional parameters, the two probing classifiers result in a total of $11.5M$ added parameters (less than 1\% of the amount of parameters of GPT-2$_{\text{XL}}$), while XLM-RoBERTa$_\text{large}$ is $558.9M$ parameters large.
Since internal representations remain the same for previous tokens during generation, \methodname can perform NER incrementally, only updating predictions for newly generated tokens, which is a novelty to the best of our knowledge.\\

\begin{table}[t]
\centering
\scalebox{0.85}{
\begin{tabular}{lccc}
\textbf{Data} & \textbf{P} & \textbf{R} & \textbf{F1} \\ \hline
\multicolumn{4}{c}{\textsc{XLM-RoBERTa$_\text{large}$}} \\
\hline
original & $84.95\%$ & $86.81\%$ & $85.87\%$ \\
generated & $83.11\%$ & $84.51\%$ & $83.81\%$ \\
$\Delta$ & \textcolor{red}{$-1.83\%$} & \textcolor{red}{$-2.30\%$} & \textcolor{red}{$-2.06\%$} \\
\hline
\multicolumn{4}{c}{\textsc{\methodnamelower (gpt-2$_\text{XL}$)}} \\
\hline
original & $82.76\%$ & $79.12\%$ & $80.90\%$ \\
generated & $86.16\%$ & $64.98\%$ & $74.09\%$ \\
$\Delta$ & \textcolor{black}{$+3.40\%$} & \textcolor{red}{$-14.14\%$} & \textcolor{red}{$-6.81\%$} \\
\hline
\multicolumn{4}{c}{\textsc{\methodnamelower (gpt-2$_\text{XL}$) trained on generated}} \\
\hline
original & $84.81\%$ & $73.63\%$ & $78.82\%$ \\
generated & $85.26\%$ & $72.05\%$ & $78.10\%$ \\
$\Delta$ & \textcolor{black}{$+0.45\%$} & \textcolor{red}{$-1.58\%$} & \textcolor{red}{$-0.72\%$} \\
\end{tabular}}
\caption{NER F1 scores for 3 approaches on our evaluation dataset. "Original" indicates the scores for the non-generated text or prompt. "Generated" indicates scores for annotations on the 100 generated tokens following the prompt.}
\label{tab:generated-ner}
\end{table}

\noindent
\textbf{Results - Accuracy.} In table \ref{tab:generated-ner} we show precision, recall and F1 scores.
For XLM-RoBERTa$_\text{large}$, we observe equal drops in precision, recall, and F1 scores of around $2\%$ on the generated text compared to the prompt. 
For \methodname trained on non-generated text, on the other hand, we measure a substantial drop in recall ($14.14\%$), while precision increases by $3.40\%$.
Further experiments\footnote{See appendices \ref{sec:appendix-generation-isolated} and \ref{sec:appendix-windowing}.} reveal that this drop in performance is caused exclusively by the span detection probe.
The results for \methodname with the span detection probe trained on generated and synthetically annotated data appear to alleviate this issue, with the F1 score only dropping by $0.72\%$ on generated text vs. the non-generated text.\\

\noindent
\textbf{Conclusion.} We show that \methodname enables vastly more efficient NER during text generation than existing approaches, increasing model parameters by less than 1\% and reducing inference speed by only around 1\% in our experiments.
We find that the attention-based span detection probing classifiers must be trained on annotated generated data in order to achieve adequate classification accuracy.
This suggests that there is a significant difference in attention weights for generated text as opposed to when non-generated text is being processed.

\section{Conclusion}

We present \methodnamepunctuation, a lightweight approach for embedding NER capabilities into decoder-only LMs without finetuning them.
We find that, except in highly data-constrained settings (such as 1-shot or 5-shot), it surpasses in-context learning in classification accuracy while being significantly more efficient.
Our approach enables efficient simultaneous text generation and NER, with only a $1\%$ reduction in token generation rate and less than $1\%$ increase in model size, paving the way for novel applications such as a significantly more efficient integration of external structured knowledge into text generation.
Lastly, we include detailed observations about the factors which influence \methodnamepunctuation's performance and provide a toolkit for training, testing, and deploying models.

\section{Outlook}

Streaming NER can provide symbolic representations of generated text at inference time in a highly efficient manner. When combined with artifacts like knowledge graphs, this could significantly accelerate applications such as real-time fact verification or retrieval-augmented generation. More broadly, exploring token classification tasks in a streaming setting could benefit safety applications—for instance, by detecting harmful outputs more rapidly—or facilitate tool integration, such as identifying mathematical symbols to trigger a calculator.

Our experiments demonstrate that reasonably accurate annotations can be achieved with our proposed method, although mention detection remains a significant bottleneck. We look forward to future research in this direction, especially regarding applications involving streaming NER and token classification in general.

\section*{Limitations}
Our findings with respect to the presented F1 scores are limited to the extent that NER is realistically modeled in the datasets used.
Specifically, inherent limitations due to autoregressivity may cause more pronounced issues in other domains or languages.
Languages other than English have not been examined in this work.
We anticipate that using our approach for languages and domains which place context relevant to entity type classification behind the mention more often than English, will cause the accuracy of predictions to suffer.
Absolute values in performance measurements referring to token generation rates will differ depending on hardware and software used.

\section*{Ethics Statement}
We acknowledge that the datasets generated for this paper were created using text generation models (GPT-2$_{\text{XL}}$), which may inadvertently produce problematic statements not reflecting our opinions.

\section*{Acknowledgements}
This work was partially supported by the German Federal Ministry of Education and Research (BMBF) as part of the Smart Data Innovation Lab (01IS19030A). The authors acknowledge support by the state of Baden-Württemberg through bwHPC. We thank Shuzhou Yuan for his feedback during editing and proofreading and our reviewers for the constructive input.

\newpage
\bibliography{anthology,custom}

\begin{thebibliography}{71}
\expandafter\ifx\csname natexlab\endcsname\relax\def\natexlab#1{#1}\fi

\bibitem[{Adi et~al.(2017)Adi, Kermany, Belinkov, Lavi, and Goldberg}]{adi2017finegrained}
Yossi Adi, Einat Kermany, Yonatan Belinkov, Ofer Lavi, and Yoav Goldberg. 2017.
\newblock \href {http://arxiv.org/abs/1608.04207} {Fine-grained analysis of sentence embeddings using auxiliary prediction tasks}.

\bibitem[{Agarap(2018)}]{relu}
Abien~Fred Agarap. 2018.
\newblock \href {http://arxiv.org/abs/1803.08375} {Deep learning using rectified linear units (relu)}.
\newblock \emph{CoRR}, abs/1803.08375.

\bibitem[{Akbik et~al.(2019)Akbik, Bergmann, Blythe, Rasul, Schweter, and Vollgraf}]{akbik2019flair}
Alan Akbik, Tanja Bergmann, Duncan Blythe, Kashif Rasul, Stefan Schweter, and Roland Vollgraf. 2019.
\newblock {FLAIR}: An easy-to-use framework for state-of-the-art {NLP}.
\newblock In \emph{{NAACL} 2019, 2019 Annual Conference of the North American Chapter of the Association for Computational Linguistics (Demonstrations)}, pages 54--59.

\bibitem[{Ashok and Lipton(2023)}]{ashok2023promptner}
Dhananjay Ashok and Zachary~C. Lipton. 2023.
\newblock \href {http://arxiv.org/abs/2305.15444} {Promptner: Prompting for named entity recognition}.

\bibitem[{Belinkov(2022)}]{belinkov-2022-probing}
Yonatan Belinkov. 2022.
\newblock \href {https://doi.org/10.1162/coli_a_00422} {Probing classifiers: Promises, shortcomings, and advances}.
\newblock \emph{Computational Linguistics}, 48(1):207--219.

\bibitem[{Belinkov and Glass(2019)}]{belinkov-glass-2019-analysis}
Yonatan Belinkov and James Glass. 2019.
\newblock \href {https://doi.org/10.1162/tacl_a_00254} {Analysis methods in neural language processing: A survey}.
\newblock \emph{Transactions of the Association for Computational Linguistics}, 7:49--72.

\bibitem[{Biderman et~al.(2023)Biderman, Schoelkopf, Anthony, Bradley, O'Brien, Hallahan, Khan, Purohit, Prashanth, Raff, Skowron, Sutawika, and van~der Wal}]{biderman2023pythia}
Stella Biderman, Hailey Schoelkopf, Quentin Anthony, Herbie Bradley, Kyle O'Brien, Eric Hallahan, Mohammad~Aflah Khan, Shivanshu Purohit, USVSN~Sai Prashanth, Edward Raff, Aviya Skowron, Lintang Sutawika, and Oskar van~der Wal. 2023.
\newblock \href {http://arxiv.org/abs/2304.01373} {Pythia: A suite for analyzing large language models across training and scaling}.

\bibitem[{Brown et~al.(2020)Brown, Mann, Ryder, Subbiah, Kaplan, Dhariwal, Neelakantan, Shyam, Sastry, Askell, Agarwal, {Herbert-Voss}, Krueger, and Henighan}]{brownLanguageModelsAre}
Tom~B Brown, Benjamin Mann, Nick Ryder, Melanie Subbiah, Jared Kaplan, Prafulla Dhariwal, Arvind Neelakantan, Pranav Shyam, Girish Sastry, Amanda Askell, Sandhini Agarwal, Ariel {Herbert-Voss}, Gretchen Krueger, and Tom Henighan. 2020.
\newblock Language {{Models}} are {{Few-Shot Learners}}.

\bibitem[{Cao et~al.(2021)Cao, Sanh, and Rush}]{cao-etal-2021-low}
Steven Cao, Victor Sanh, and Alexander Rush. 2021.
\newblock \href {https://doi.org/10.18653/v1/2021.naacl-main.74} {Low-complexity probing via finding subnetworks}.
\newblock In \emph{Proceedings of the 2021 Conference of the North American Chapter of the Association for Computational Linguistics: Human Language Technologies}, pages 960--966, Online. Association for Computational Linguistics.

\bibitem[{Chen et~al.(2023{\natexlab{a}})Chen, Pasupat, Singh, Lee, and Guu}]{chen2023purr}
Anthony Chen, Panupong Pasupat, Sameer Singh, Hongrae Lee, and Kelvin Guu. 2023{\natexlab{a}}.
\newblock \href {http://arxiv.org/abs/2305.14908} {Purr: Efficiently editing language model hallucinations by denoising language model corruptions}.

\bibitem[{Chen et~al.(2023{\natexlab{b}})Chen, Lu, Lin, Lou, Jia, Dai, Wu, Cao, Han, and Sun}]{chen-etal-2023-learning}
Jiawei Chen, Yaojie Lu, Hongyu Lin, Jie Lou, Wei Jia, Dai Dai, Hua Wu, Boxi Cao, Xianpei Han, and Le~Sun. 2023{\natexlab{b}}.
\newblock \href {https://doi.org/10.18653/v1/2023.acl-long.764} {Learning in-context learning for named entity recognition}.
\newblock In \emph{Proceedings of the 61st Annual Meeting of the Association for Computational Linguistics (Volume 1: Long Papers)}, pages 13661--13675, Toronto, Canada. Association for Computational Linguistics.

\bibitem[{Chowdhery et~al.(2022)Chowdhery, Narang, Devlin, Bosma, Mishra, Roberts, Barham, Chung, Sutton, and Gehrmann~et al.}]{chowdheryPaLMScalingLanguage2022}
Aakanksha Chowdhery, Sharan Narang, Jacob Devlin, Maarten Bosma, Gaurav Mishra, Adam Roberts, Paul Barham, Hyung~Won Chung, Charles Sutton, and Sebastian Gehrmann~et al. 2022.
\newblock \href {https://doi.org/10.48550/arXiv.2204.02311} {{{PaLM}}: {{Scaling Language Modeling}} with {{Pathways}}}.

\bibitem[{Clark et~al.(2019)Clark, Khandelwal, Levy, and Manning}]{clark-etal-2019-bert}
Kevin Clark, Urvashi Khandelwal, Omer Levy, and Christopher~D. Manning. 2019.
\newblock \href {https://doi.org/10.18653/v1/W19-4828} {What does {BERT} look at? an analysis of {BERT}{'}s attention}.
\newblock In \emph{Proceedings of the 2019 ACL Workshop BlackboxNLP: Analyzing and Interpreting Neural Networks for NLP}, pages 276--286, Florence, Italy. Association for Computational Linguistics.

\bibitem[{Derczynski et~al.(2017)Derczynski, Nichols, van Erp, and Limsopatham}]{derczynski-etal-2017-results}
Leon Derczynski, Eric Nichols, Marieke van Erp, and Nut Limsopatham. 2017.
\newblock \href {https://doi.org/10.18653/v1/W17-4418} {Results of the {WNUT}2017 shared task on novel and emerging entity recognition}.
\newblock In \emph{Proceedings of the 3rd Workshop on Noisy User-generated Text}, pages 140--147, Copenhagen, Denmark. Association for Computational Linguistics.

\bibitem[{Dhuliawala et~al.(2023)Dhuliawala, Komeili, Xu, Raileanu, Li, Celikyilmaz, and Weston}]{dhuliawalaChainofVerificationReducesHallucination2023}
Shehzaad Dhuliawala, Mojtaba Komeili, Jing Xu, Roberta Raileanu, Xian Li, Asli Celikyilmaz, and Jason Weston. 2023.
\newblock \href {https://doi.org/10.48550/arXiv.2309.11495} {Chain-of-{{Verification Reduces Hallucination}} in {{Large Language Models}}}.

\bibitem[{Dubey et~al.(2024)Dubey, Jauhri, Pandey, Kadian, Al-Dahle, Letman, Mathur, Schelten, Yang, and et~al.}]{dubey2024llama3herdmodels}
Abhimanyu Dubey, Abhinav Jauhri, Abhinav Pandey, Abhishek Kadian, Ahmad Al-Dahle, Aiesha Letman, Akhil Mathur, Alan Schelten, Amy Yang, and Angela~Fan et~al. 2024.
\newblock \href {http://arxiv.org/abs/2407.21783} {The llama 3 herd of models}.

\bibitem[{Epure and Hennequin(2022)}]{epure-hennequin-2022-probing}
Elena~V. Epure and Romain Hennequin. 2022.
\newblock \href {https://aclanthology.org/2022.lrec-1.151} {Probing pre-trained auto-regressive language models for named entity typing and recognition}.
\newblock In \emph{Proceedings of the Thirteenth Language Resources and Evaluation Conference}, pages 1408--1417, Marseille, France. European Language Resources Association.

\bibitem[{Ettinger et~al.(2016)Ettinger, Elgohary, and Resnik}]{ettinger-etal-2016-probing}
Allyson Ettinger, Ahmed Elgohary, and Philip Resnik. 2016.
\newblock \href {https://doi.org/10.18653/v1/W16-2524} {Probing for semantic evidence of composition by means of simple classification tasks}.
\newblock In \emph{Proceedings of the 1st Workshop on Evaluating Vector-Space Representations for {NLP}}, pages 134--139, Berlin, Germany. Association for Computational Linguistics.

\bibitem[{Fu et~al.(2021)Fu, Huang, and Liu}]{fu-etal-2021-spanner}
Jinlan Fu, Xuanjing Huang, and Pengfei Liu. 2021.
\newblock \href {https://doi.org/10.18653/v1/2021.acl-long.558} {{S}pan{NER}: Named entity re-/recognition as span prediction}.
\newblock In \emph{Proceedings of the 59th Annual Meeting of the Association for Computational Linguistics and the 11th International Joint Conference on Natural Language Processing (Volume 1: Long Papers)}, pages 7183--7195, Online. Association for Computational Linguistics.

\bibitem[{Gao et~al.(2024)Gao, Xiong, Gao, Jia, Pan, Bi, Dai, Sun, Guo, Wang, and Wang}]{gaoRetrievalAugmentedGenerationLarge2024}
Yunfan Gao, Yun Xiong, Xinyu Gao, Kangxiang Jia, Jinliu Pan, Yuxi Bi, Yi~Dai, Jiawei Sun, Qianyu Guo, Meng Wang, and Haofen Wang. 2024.
\newblock \href {https://doi.org/10.48550/arXiv.2312.10997} {Retrieval-{{Augmented Generation}} for {{Large Language Models}}: {{A Survey}}}.

\bibitem[{Geva et~al.(2023)Geva, Bastings, Filippova, and Globerson}]{gevaDissectingRecallFactual2023a}
Mor Geva, Jasmijn Bastings, Katja Filippova, and Amir Globerson. 2023.
\newblock \href {https://doi.org/10.48550/arXiv.2304.14767} {Dissecting {{Recall}} of {{Factual Associations}} in {{Auto-Regressive Language Models}}}.

\bibitem[{Ghandeharioun et~al.(2024)Ghandeharioun, Caciularu, Pearce, Dixon, and Geva}]{ghandehariounPatchscopesUnifyingFramework2024}
Asma Ghandeharioun, Avi Caciularu, Adam Pearce, Lucas Dixon, and Mor Geva. 2024.
\newblock \href {https://doi.org/10.48550/arXiv.2401.06102} {Patchscopes: {{A Unifying Framework}} for {{Inspecting Hidden Representations}} of {{Language Models}}}.

\bibitem[{Goodfellow et~al.(2015)Goodfellow, Mirza, Xiao, Courville, and Bengio}]{goodfellowEmpiricalInvestigationCatastrophic2015}
Ian~J. Goodfellow, Mehdi Mirza, Da~Xiao, Aaron Courville, and Yoshua Bengio. 2015.
\newblock \href {http://arxiv.org/abs/1312.6211} {An {{Empirical Investigation}} of {{Catastrophic Forgetting}} in {{Gradient-Based Neural Networks}}}.

\bibitem[{Goyal et~al.(2017)Goyal, Dollár, Girshick, Noordhuis, Wesolowski, Kyrola, Tulloch, Jia, and He}]{goyal_accurate_2017}
Priya Goyal, Piotr Dollár, Ross Girshick, Pieter Noordhuis, Lukasz Wesolowski, Aapo Kyrola, Andrew Tulloch, Yangqing Jia, and Kaiming He. 2017.
\newblock \href {https://arxiv.org/abs/1706.02677v1} {Accurate, {Large} {Minibatch} {SGD}: {Training} {ImageNet} in 1 {Hour}}.

\bibitem[{Guo et~al.(2023)Guo, Li, Jin, Liu, Zeng, Liu, Li, Yang, Bai, Guo, and Cheng}]{guo2023retrievalaugmented}
Yucan Guo, Zixuan Li, Xiaolong Jin, Yantao Liu, Yutao Zeng, Wenxuan Liu, Xiang Li, Pan Yang, Long Bai, Jiafeng Guo, and Xueqi Cheng. 2023.
\newblock \href {http://arxiv.org/abs/2311.02962} {Retrieval-augmented code generation for universal information extraction}.

\bibitem[{Guu et~al.(2020)Guu, Lee, Tung, Pasupat, and Chang}]{pmlr-v119-guu20a}
Kelvin Guu, Kenton Lee, Zora Tung, Panupong Pasupat, and Mingwei Chang. 2020.
\newblock \href {https://proceedings.mlr.press/v119/guu20a.html} {Retrieval augmented language model pre-training}.
\newblock In \emph{Proceedings of the 37th International Conference on Machine Learning}, volume 119 of \emph{Proceedings of Machine Learning Research}, pages 3929--3938. PMLR.

\bibitem[{Hernandez et~al.(2023)Hernandez, Sharma, Haklay, Meng, Wattenberg, Andreas, Belinkov, and Bau}]{hernandezLinearityRelationDecoding2023a}
Evan Hernandez, Arnab~Sen Sharma, Tal Haklay, Kevin Meng, Martin Wattenberg, Jacob Andreas, Yonatan Belinkov, and David Bau. 2023.
\newblock \href {https://doi.org/10.48550/arXiv.2308.09124} {Linearity of {{Relation Decoding}} in {{Transformer Language Models}}}.

\bibitem[{Hovy et~al.(2006)Hovy, Marcus, Palmer, Ramshaw, and Weischedel}]{hovy-etal-2006-ontonotes}
Eduard Hovy, Mitchell Marcus, Martha Palmer, Lance Ramshaw, and Ralph Weischedel. 2006.
\newblock \href {https://aclanthology.org/N06-2015} {{O}nto{N}otes: The 90{\%} solution}.
\newblock In \emph{Proceedings of the Human Language Technology Conference of the {NAACL}, Companion Volume: Short Papers}, pages 57--60, New York City, USA. Association for Computational Linguistics.

\bibitem[{Htut et~al.(2019)Htut, Phang, Bordia, and Bowman}]{htutAttentionHeadsBERT2019}
Phu~Mon Htut, Jason Phang, Shikha Bordia, and Samuel~R. Bowman. 2019.
\newblock \href {https://doi.org/10.48550/arXiv.1911.12246} {Do {{Attention Heads}} in {{BERT Track Syntactic Dependencies}}?}

\bibitem[{Hu et~al.(2021)Hu, Shen, Wallis, Allen-Zhu, Li, Wang, Wang, and Chen}]{hu2021lora}
Edward~J. Hu, Yelong Shen, Phillip Wallis, Zeyuan Allen-Zhu, Yuanzhi Li, Shean Wang, Lu~Wang, and Weizhu Chen. 2021.
\newblock \href {http://arxiv.org/abs/2106.09685} {Lora: Low-rank adaptation of large language models}.

\bibitem[{Josifoski et~al.(2022)Josifoski, De~Cao, Peyrard, Petroni, and West}]{josifoski-etal-2022-genie}
Martin Josifoski, Nicola De~Cao, Maxime Peyrard, Fabio Petroni, and Robert West. 2022.
\newblock \href {https://doi.org/10.18653/v1/2022.naacl-main.342} {{G}en{IE}: Generative information extraction}.
\newblock In \emph{Proceedings of the 2022 Conference of the North American Chapter of the Association for Computational Linguistics: Human Language Technologies}, pages 4626--4643, Seattle, United States. Association for Computational Linguistics.

\bibitem[{Keskar et~al.(2019)Keskar, McCann, Varshney, Xiong, and Socher}]{keskarCTRLConditionalTransformer2019}
Nitish~Shirish Keskar, Bryan McCann, Lav~R. Varshney, Caiming Xiong, and Richard Socher. 2019.
\newblock \href {https://doi.org/10.48550/arXiv.1909.05858} {{{CTRL}}: {{A Conditional Transformer Language Model}} for {{Controllable Generation}}}.

\bibitem[{Lewis et~al.(2020)Lewis, Perez, Piktus, Petroni, Karpukhin, Goyal, K\"{u}ttler, Lewis, Yih, Rockt\"{a}schel, Riedel, and Kiela}]{NEURIPS2020_6b493230}
Patrick Lewis, Ethan Perez, Aleksandra Piktus, Fabio Petroni, Vladimir Karpukhin, Naman Goyal, Heinrich K\"{u}ttler, Mike Lewis, Wen-tau Yih, Tim Rockt\"{a}schel, Sebastian Riedel, and Douwe Kiela. 2020.
\newblock \href {https://proceedings.neurips.cc/paper_files/paper/2020/file/6b493230205f780e1bc26945df7481e5-Paper.pdf} {Retrieval-augmented generation for knowledge-intensive nlp tasks}.
\newblock In \emph{Advances in Neural Information Processing Systems}, volume~33, pages 9459--9474. Curran Associates, Inc.

\bibitem[{Li et~al.(2016)Li, Sun, Johnson, Sciaky, Wei, Leaman, Davis, Mattingly, Wiegers, and Lu}]{liBioCreativeCDRTask2016a}
Jiao Li, Yueping Sun, Robin~J. Johnson, Daniela Sciaky, Chih-Hsuan Wei, Robert Leaman, Allan~Peter Davis, Carolyn~J. Mattingly, Thomas~C. Wiegers, and Zhiyong Lu. 2016.
\newblock \href {https://doi.org/10.1093/database/baw068} {{{BioCreative V CDR}} task corpus: A resource for chemical disease relation extraction}.
\newblock \emph{Database: The Journal of Biological Databases and Curation}, 2016:baw068.

\bibitem[{Li et~al.(2022)Li, Sun, Han, and Li}]{liSurveyDeepLearning2022}
Jing Li, Aixin Sun, Jianglei Han, and Chenliang Li. 2022.
\newblock \href {https://doi.org/10.1109/TKDE.2020.2981314} {A {{Survey}} on {{Deep Learning}} for {{Named Entity Recognition}}}.
\newblock \emph{IEEE Transactions on Knowledge and Data Engineering}, 34(1):50--70.

\bibitem[{Li et~al.(2023)Li, Sun, Tang, Yan, Wu, Huang, and Qiu}]{li-etal-2023-codeie}
Peng Li, Tianxiang Sun, Qiong Tang, Hang Yan, Yuanbin Wu, Xuanjing Huang, and Xipeng Qiu. 2023.
\newblock \href {https://doi.org/10.18653/v1/2023.acl-long.855} {{C}ode{IE}: Large code generation models are better few-shot information extractors}.
\newblock In \emph{Proceedings of the 61st Annual Meeting of the Association for Computational Linguistics (Volume 1: Long Papers)}, pages 15339--15353, Toronto, Canada. Association for Computational Linguistics.

\bibitem[{Loshchilov and Hutter(2019)}]{loshchilov_decoupled_2019}
Ilya Loshchilov and Frank Hutter. 2019.
\newblock Decoupled {Weight} {Decay} {Regularization}.
\newblock In \emph{Proceedings of the International Conference on Learning Representations 2019}, page~18.

\bibitem[{Lu et~al.(2022)Lu, Liu, Dai, Xiao, Lin, Han, Sun, and Wu}]{lu-etal-2022-unified}
Yaojie Lu, Qing Liu, Dai Dai, Xinyan Xiao, Hongyu Lin, Xianpei Han, Le~Sun, and Hua Wu. 2022.
\newblock \href {https://doi.org/10.18653/v1/2022.acl-long.395} {Unified structure generation for universal information extraction}.
\newblock In \emph{Proceedings of the 60th Annual Meeting of the Association for Computational Linguistics (Volume 1: Long Papers)}, pages 5755--5772, Dublin, Ireland. Association for Computational Linguistics.

\bibitem[{Luo et~al.(2020)Luo, Xiao, and Zhao}]{Luo_Xiao_Zhao_2020}
Ying Luo, Fengshun Xiao, and Hai Zhao. 2020.
\newblock \href {https://doi.org/10.1609/aaai.v34i05.6363} {Hierarchical contextualized representation for named entity recognition}.
\newblock \emph{Proceedings of the AAAI Conference on Artificial Intelligence}, 34(05):8441--8448.

\bibitem[{Mare{\v{c}}ek and Rosa(2019)}]{marecek-rosa-2019-balustrades}
David Mare{\v{c}}ek and Rudolf Rosa. 2019.
\newblock \href {https://doi.org/10.18653/v1/W19-4827} {From balustrades to pierre vinken: Looking for syntax in transformer self-attentions}.
\newblock In \emph{Proceedings of the 2019 ACL Workshop BlackboxNLP: Analyzing and Interpreting Neural Networks for NLP}, pages 263--275, Florence, Italy. Association for Computational Linguistics.

\bibitem[{Meng et~al.(2022)Meng, Bau, Andonian, and Belinkov}]{mengLocatingEditingFactual2022c}
Kevin Meng, David Bau, Alex~J. Andonian, and Yonatan Belinkov. 2022.
\newblock Locating and {{Editing Factual Associations}} in {{GPT}}.
\newblock In \emph{Advances in {{Neural Information Processing Systems}}}.

\bibitem[{Nakayama(2018)}]{seqeval}
Hiroki Nakayama. 2018.
\newblock \href {https://github.com/chakki-works/seqeval} {{seqeval}: A python framework for sequence labeling evaluation}.
\newblock Software available from https://github.com/chakki-works/seqeval.

\bibitem[{Nie et~al.(2024)Nie, Shao, and Wang}]{nie-etal-2024-know-adapter}
Binling Nie, Yiming Shao, and Yigang Wang. 2024.
\newblock \href {https://aclanthology.org/2024.lrec-main.854} {Know-adapter: Towards knowledge-aware parameter-efficient transfer learning for few-shot named entity recognition}.
\newblock In \emph{Proceedings of the 2024 Joint International Conference on Computational Linguistics, Language Resources and Evaluation (LREC-COLING 2024)}, pages 9777--9786, Torino, Italia. ELRA and ICCL.

\bibitem[{Pimentel et~al.(2020)Pimentel, Valvoda, Maudslay, Zmigrod, Williams, and Cotterell}]{pimentel-etal-2020-information}
Tiago Pimentel, Josef Valvoda, Rowan~Hall Maudslay, Ran Zmigrod, Adina Williams, and Ryan Cotterell. 2020.
\newblock \href {https://doi.org/10.18653/v1/2020.acl-main.420} {Information-theoretic probing for linguistic structure}.
\newblock In \emph{Proceedings of the 58th Annual Meeting of the Association for Computational Linguistics}, pages 4609--4622, Online. Association for Computational Linguistics.

\bibitem[{Radford et~al.(2019)Radford, Wu, Child, Luan, Amodei, and Sutskever}]{radford2019language}
Alec Radford, Jeff Wu, Rewon Child, David Luan, Dario Amodei, and Ilya Sutskever. 2019.
\newblock Language models are unsupervised multitask learners.

\bibitem[{Raganato and Tiedemann(2018)}]{raganato-tiedemann-2018-analysis}
Alessandro Raganato and J{\"o}rg Tiedemann. 2018.
\newblock \href {https://doi.org/10.18653/v1/W18-5431} {An analysis of encoder representations in transformer-based machine translation}.
\newblock In \emph{Proceedings of the 2018 {EMNLP} Workshop {B}lackbox{NLP}: Analyzing and Interpreting Neural Networks for {NLP}}, pages 287--297, Brussels, Belgium. Association for Computational Linguistics.

\bibitem[{Ram et~al.(2023)Ram, Levine, Dalmedigos, Muhlgay, Shashua, Leyton-Brown, and Shoham}]{ram-etal-2023-context}
Ori Ram, Yoav Levine, Itay Dalmedigos, Dor Muhlgay, Amnon Shashua, Kevin Leyton-Brown, and Yoav Shoham. 2023.
\newblock \href {https://doi.org/10.1162/tacl_a_00605} {In-context retrieval-augmented language models}.
\newblock \emph{Transactions of the Association for Computational Linguistics}, 11:1316--1331.

\bibitem[{Ruder et~al.(2019)Ruder, S{\o}gaard, and Vuli{\'c}}]{ruder-etal-2019-unsupervised}
Sebastian Ruder, Anders S{\o}gaard, and Ivan Vuli{\'c}. 2019.
\newblock \href {https://doi.org/10.18653/v1/P19-4007} {Unsupervised cross-lingual representation learning}.
\newblock In \emph{Proceedings of the 57th Annual Meeting of the Association for Computational Linguistics: Tutorial Abstracts}, pages 31--38, Florence, Italy. Association for Computational Linguistics.

\bibitem[{Scao et~al.(2022)Scao, Fan, Akiki, Pavlick, Ili{\'c}, Hesslow, Castagn{\'e}, Luccioni, Yvon, and Gall{\'e}~et al.}]{workshopBLOOM176BParameterOpenAccess2022}
Teven~Le Scao, Angela Fan, Christopher Akiki, Ellie Pavlick, Suzana Ili{\'c}, Daniel Hesslow, Roman Castagn{\'e}, Alexandra~Sasha Luccioni, Fran{\c c}ois Yvon, and Matthias Gall{\'e}~et al. 2022.
\newblock \href {https://doi.org/10.48550/arXiv.2211.05100} {{{BLOOM}}: {{A 176B-Parameter Open-Access Multilingual Language Model}}}.

\bibitem[{Schick et~al.(2023)Schick, {Dwivedi-Yu}, Dess{\`i}, Raileanu, Lomeli, Zettlemoyer, Cancedda, and Scialom}]{schickToolformerLanguageModels2023a}
Timo Schick, Jane {Dwivedi-Yu}, Roberto Dess{\`i}, Roberta Raileanu, Maria Lomeli, Luke Zettlemoyer, Nicola Cancedda, and Thomas Scialom. 2023.
\newblock \href {https://doi.org/10.48550/arXiv.2302.04761} {Toolformer: {{Language Models Can Teach Themselves}} to {{Use Tools}}}.

\bibitem[{Schouten et~al.(2022)Schouten, Bloem, and Vossen}]{schouten-etal-2022-probing}
Stefan Schouten, Peter Bloem, and Piek Vossen. 2022.
\newblock \href {https://doi.org/10.18653/v1/2022.blackboxnlp-1.32} {Probing the representations of named entities in transformer-based language models}.
\newblock In \emph{Proceedings of the Fifth BlackboxNLP Workshop on Analyzing and Interpreting Neural Networks for NLP}, pages 384--393, Abu Dhabi, United Arab Emirates (Hybrid). Association for Computational Linguistics.

\bibitem[{Shen et~al.(2023)Shen, Tan, Wu, Zhang, Zhang, Xi, Lu, and Zhuang}]{shen-etal-2023-promptner}
Yongliang Shen, Zeqi Tan, Shuhui Wu, Wenqi Zhang, Rongsheng Zhang, Yadong Xi, Weiming Lu, and Yueting Zhuang. 2023.
\newblock \href {https://doi.org/10.18653/v1/2023.acl-long.698} {{P}rompt{NER}: Prompt locating and typing for named entity recognition}.
\newblock In \emph{Proceedings of the 61st Annual Meeting of the Association for Computational Linguistics (Volume 1: Long Papers)}, pages 12492--12507, Toronto, Canada. Association for Computational Linguistics.

\bibitem[{Shi et~al.(2023)Shi, Min, Yasunaga, Seo, James, Lewis, Zettlemoyer, and Yih}]{shiREPLUGRetrievalAugmentedBlackBox2023}
Weijia Shi, Sewon Min, Michihiro Yasunaga, Minjoon Seo, Rich James, Mike Lewis, Luke Zettlemoyer, and Wen-tau Yih. 2023.
\newblock \href {https://doi.org/10.48550/arXiv.2301.12652} {{{REPLUG}}: {{Retrieval-Augmented Black-Box Language Models}}}.

\bibitem[{Shi et~al.(2016)Shi, Padhi, and Knight}]{shi-etal-2016-string}
Xing Shi, Inkit Padhi, and Kevin Knight. 2016.
\newblock \href {https://doi.org/10.18653/v1/D16-1159} {Does string-based neural {MT} learn source syntax?}
\newblock In \emph{Proceedings of the 2016 Conference on Empirical Methods in Natural Language Processing}, pages 1526--1534, Austin, Texas. Association for Computational Linguistics.

\bibitem[{Tan et~al.(2021)Tan, Shen, Zhang, Lu, and Zhuang}]{ijcai2021p542}
Zeqi Tan, Yongliang Shen, Shuai Zhang, Weiming Lu, and Yueting Zhuang. 2021.
\newblock \href {https://doi.org/10.24963/ijcai.2021/542} {A sequence-to-set network for nested named entity recognition}.
\newblock In \emph{Proceedings of the Thirtieth International Joint Conference on Artificial Intelligence, {IJCAI-21}}, pages 3936--3942. International Joint Conferences on Artificial Intelligence Organization.
\newblock Main Track.

\bibitem[{Tenney et~al.(2019)Tenney, Xia, Chen, Wang, Poliak, McCoy, Kim, Durme, Bowman, Das, and Pavlick}]{tenney2018what}
Ian Tenney, Patrick Xia, Berlin Chen, Alex Wang, Adam Poliak, R~Thomas McCoy, Najoung Kim, Benjamin~Van Durme, Sam Bowman, Dipanjan Das, and Ellie Pavlick. 2019.
\newblock \href {https://openreview.net/forum?id=SJzSgnRcKX} {What do you learn from context? probing for sentence structure in contextualized word representations}.
\newblock In \emph{International Conference on Learning Representations}.

\bibitem[{Tjong Kim~Sang and De~Meulder(2003{\natexlab{a}})}]{tjong-kim-sang-de-meulder-2003-introduction}
Erik~F. Tjong Kim~Sang and Fien De~Meulder. 2003{\natexlab{a}}.
\newblock \href {https://aclanthology.org/W03-0419} {Introduction to the {C}o{NLL}-2003 shared task: Language-independent named entity recognition}.
\newblock In \emph{Proceedings of the Seventh Conference on Natural Language Learning at {HLT}-{NAACL} 2003}, pages 142--147.

\bibitem[{Tjong Kim~Sang and De~Meulder(2003{\natexlab{b}})}]{tjongkimsang2003conll}
Erik~F. Tjong Kim~Sang and Fien De~Meulder. 2003{\natexlab{b}}.
\newblock Introduction to the conll-2003 shared task: Language-independent named entity recognition.
\newblock In \emph{Proceedings of CoNLL-2003}, pages 142--147. Edmonton, Canada.

\bibitem[{Touvron et~al.(2023)Touvron, Lavril, Izacard, Martinet, Lachaux, Lacroix, Rozi{\`e}re, Goyal, Hambro, Azhar, Rodriguez, Joulin, Grave, and Lample}]{touvronLLaMAOpenEfficient2023}
Hugo Touvron, Thibaut Lavril, Gautier Izacard, Xavier Martinet, Marie-Anne Lachaux, Timoth{\'e}e Lacroix, Baptiste Rozi{\`e}re, Naman Goyal, Eric Hambro, Faisal Azhar, Aurelien Rodriguez, Armand Joulin, Edouard Grave, and Guillaume Lample. 2023.
\newblock \href {https://doi.org/10.48550/arXiv.2302.13971} {{{LLaMA}}: {{Open}} and {{Efficient Foundation Language Models}}}.

\bibitem[{Wang and Komatsuzaki(2021)}]{gpt-j}
Ben Wang and Aran Komatsuzaki. 2021.
\newblock {GPT-J-6B: A 6 Billion Parameter Autoregressive Language Model}.
\newblock \url{https://github.com/kingoflolz/mesh-transformer-jax}.

\bibitem[{Wang et~al.(2023{\natexlab{a}})Wang, Li, Dai, Chen, Zhou, Meng, Zhou, and Sun}]{wang-etal-2023-label}
Lean Wang, Lei Li, Damai Dai, Deli Chen, Hao Zhou, Fandong Meng, Jie Zhou, and Xu~Sun. 2023{\natexlab{a}}.
\newblock \href {https://doi.org/10.18653/v1/2023.emnlp-main.609} {Label words are anchors: An information flow perspective for understanding in-context learning}.
\newblock In \emph{Proceedings of the 2023 Conference on Empirical Methods in Natural Language Processing}, pages 9840--9855, Singapore. Association for Computational Linguistics.

\bibitem[{Wang et~al.(2023{\natexlab{b}})Wang, Sun, Li, Ouyang, Wu, Zhang, Li, and Wang}]{wang2023gpt}
Shuhe Wang, Xiaofei Sun, Xiaoya Li, Rongbin Ouyang, Fei Wu, Tianwei Zhang, Jiwei Li, and Guoyin Wang. 2023{\natexlab{b}}.
\newblock Gpt-ner: Named entity recognition via large language models.
\newblock \emph{arXiv preprint arXiv:2304.10428}.

\bibitem[{Wang et~al.(2021{\natexlab{a}})Wang, Jiang, Bach, Wang, Huang, Huang, and Tu}]{wang-etal-2021-automated}
Xinyu Wang, Yong Jiang, Nguyen Bach, Tao Wang, Zhongqiang Huang, Fei Huang, and Kewei Tu. 2021{\natexlab{a}}.
\newblock \href {https://doi.org/10.18653/v1/2021.acl-long.206} {Automated concatenation of embeddings for structured prediction}.
\newblock In \emph{Proceedings of the 59th Annual Meeting of the Association for Computational Linguistics and the 11th International Joint Conference on Natural Language Processing (Volume 1: Long Papers)}, pages 2643--2660, Online. Association for Computational Linguistics.

\bibitem[{Wang et~al.(2021{\natexlab{b}})Wang, Jiang, Bach, Wang, Huang, Huang, and Tu}]{wang-etal-2021-improving}
Xinyu Wang, Yong Jiang, Nguyen Bach, Tao Wang, Zhongqiang Huang, Fei Huang, and Kewei Tu. 2021{\natexlab{b}}.
\newblock \href {https://doi.org/10.18653/v1/2021.acl-long.142} {Improving named entity recognition by external context retrieving and cooperative learning}.
\newblock In \emph{Proceedings of the 59th Annual Meeting of the Association for Computational Linguistics and the 11th International Joint Conference on Natural Language Processing (Volume 1: Long Papers)}, pages 1800--1812, Online. Association for Computational Linguistics.

\bibitem[{Wolf et~al.(2020)Wolf, Debut, Sanh, Chaumond, Delangue, Moi, Cistac, Rault, Louf, Funtowicz, Davison, Shleifer, von Platen, Ma, Jernite, Plu, Xu, Le~Scao, Gugger, Drame, Lhoest, and Rush}]{wolf-etal-2020-transformers}
Thomas Wolf, Lysandre Debut, Victor Sanh, Julien Chaumond, Clement Delangue, Anthony Moi, Pierric Cistac, Tim Rault, Remi Louf, Morgan Funtowicz, Joe Davison, Sam Shleifer, Patrick von Platen, Clara Ma, Yacine Jernite, Julien Plu, Canwen Xu, Teven Le~Scao, Sylvain Gugger, Mariama Drame, Quentin Lhoest, and Alexander Rush. 2020.
\newblock \href {https://doi.org/10.18653/v1/2020.emnlp-demos.6} {Transformers: State-of-the-art natural language processing}.
\newblock In \emph{Proceedings of the 2020 Conference on Empirical Methods in Natural Language Processing: System Demonstrations}, pages 38--45, Online. Association for Computational Linguistics.

\bibitem[{Yan et~al.(2021)Yan, Gui, Dai, Guo, Zhang, and Qiu}]{yan-etal-2021-unified-generative}
Hang Yan, Tao Gui, Junqi Dai, Qipeng Guo, Zheng Zhang, and Xipeng Qiu. 2021.
\newblock \href {https://doi.org/10.18653/v1/2021.acl-long.451} {A unified generative framework for various {NER} subtasks}.
\newblock In \emph{Proceedings of the 59th Annual Meeting of the Association for Computational Linguistics and the 11th International Joint Conference on Natural Language Processing (Volume 1: Long Papers)}, pages 5808--5822, Online. Association for Computational Linguistics.

\bibitem[{Ye et~al.(2022)Ye, Lin, Li, and Sun}]{ye-etal-2022-packed}
Deming Ye, Yankai Lin, Peng Li, and Maosong Sun. 2022.
\newblock \href {https://doi.org/10.18653/v1/2022.acl-long.337} {Packed levitated marker for entity and relation extraction}.
\newblock In \emph{Proceedings of the 60th Annual Meeting of the Association for Computational Linguistics (Volume 1: Long Papers)}, pages 4904--4917, Dublin, Ireland. Association for Computational Linguistics.

\bibitem[{Zhang(2023)}]{Zhang2023GraphToolFormerTE}
Jiawei Zhang. 2023.
\newblock Graph-toolformer: To empower llms with graph reasoning ability via prompt augmented by chatgpt.
\newblock \emph{ArXiv}, abs/2304.11116.

\bibitem[{Zhang et~al.(2023{\natexlab{a}})Zhang, Cheng, Gao, and Poon}]{zhang2023optimizing}
Sheng Zhang, Hao Cheng, Jianfeng Gao, and Hoifung Poon. 2023{\natexlab{a}}.
\newblock \href {http://arxiv.org/abs/2208.14565} {Optimizing bi-encoder for named entity recognition via contrastive learning}.

\bibitem[{Zhang et~al.(2022)Zhang, Roller, Goyal, Artetxe, Chen, Chen, Dewan, Diab, Li, Lin, Mihaylov, Ott, Shleifer, Shuster, Simig, Koura, Sridhar, Wang, and Zettlemoyer}]{zhangOPTOpenPretrained2022}
Susan Zhang, Stephen Roller, Naman Goyal, Mikel Artetxe, Moya Chen, Shuohui Chen, Christopher Dewan, Mona Diab, Xian Li, Xi~Victoria Lin, Todor Mihaylov, Myle Ott, Sam Shleifer, Kurt Shuster, Daniel Simig, Punit~Singh Koura, Anjali Sridhar, Tianlu Wang, and Luke Zettlemoyer. 2022.
\newblock \href {https://doi.org/10.48550/arXiv.2205.01068} {{{OPT}}: {{Open Pre-trained Transformer Language Models}}}.

\bibitem[{Zhang et~al.(2023{\natexlab{b}})Zhang, Li, Cui, Cai, Liu, Fu, Huang, Zhao, Zhang, Chen, Wang, Luu, Bi, Shi, and Shi}]{zhang2023sirens}
Yue Zhang, Yafu Li, Leyang Cui, Deng Cai, Lemao Liu, Tingchen Fu, Xinting Huang, Enbo Zhao, Yu~Zhang, Yulong Chen, Longyue Wang, Anh~Tuan Luu, Wei Bi, Freda Shi, and Shuming Shi. 2023{\natexlab{b}}.
\newblock \href {http://arxiv.org/abs/2309.01219} {Siren's song in the ai ocean: A survey on hallucination in large language models}.

\end{thebibliography}

\appendix

\begin{figure*}[!htb]
    \centering
    \scalebox{0.5}{
    \includegraphics{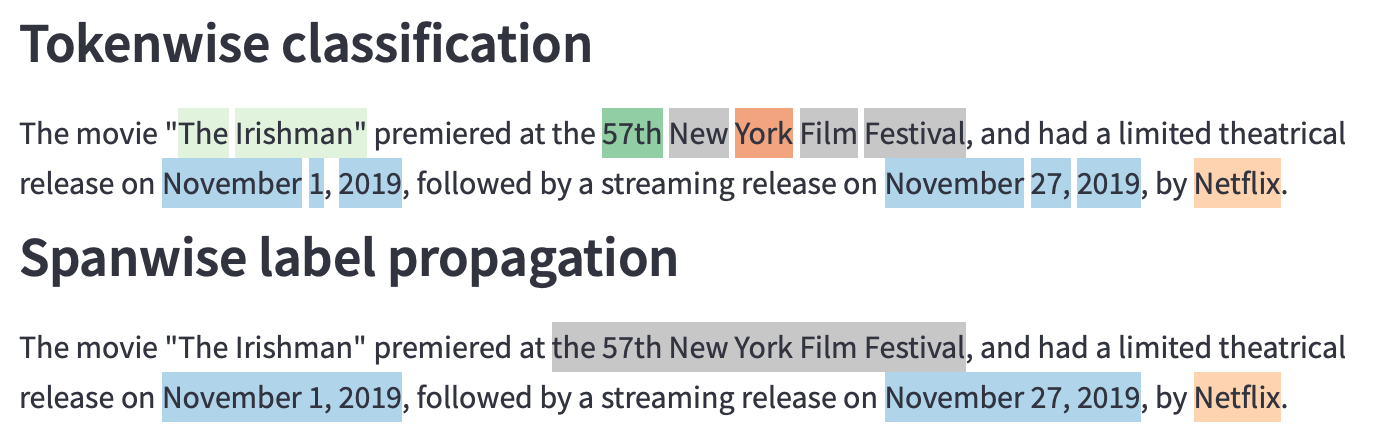}
    }
    \caption{Example NER output of \methodname trained on Ontonotes5 and GPT-2$_\text{XL}$. Colors indicate different predicted entity types. The example illustrates both a failure case due to missed span detection causing correct type predictions to be discarded (\textit{``The Irishman''}, type: ``WORK OF ART''), as well as spanwise label propagation applying the correct entity type (``EVENT'') to a multi-token span based on the type predicted for the last token (\textit{``The 57th New York Film Festival''}).}
    \label{fig:example-output}
\end{figure*}

\begin{figure*}[!htb]
    \centering
    \scalebox{0.5}{
    \includegraphics{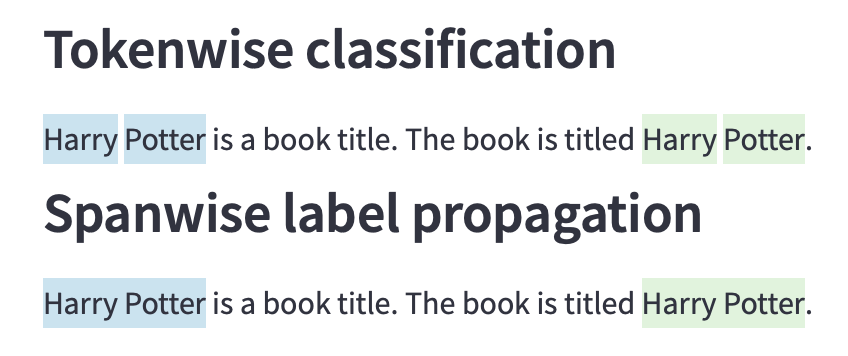}
    }
    \caption{Example NER output of \methodname trained on Ontonotes5 and GPT-2$_\text{XL}$. Colors indicate different predicted entity types (blue: ``PERSON'', green: ``WORK OF ART''). The example illustrates an inherent limitation of our approach due to autoregressivity, where the first mention of \textit{``Harry Potter''} is misclassified as ``PERSON''. The second mention is correctly classified as ``WORK OF ART'' since the required context precedes the entity mention.}
    \label{fig:example-output-hp}
\end{figure*}
\section{Implementation Details}
\label{sec:appendix-implementation-details}

\noindent
\textbf{Data and Models.} All experiments are performed using the datasets CoNLL2003 \cite{tjong-kim-sang-de-meulder-2003-introduction} and Ontonotes5 \cite{hovy-etal-2006-ontonotes} and 7 LMs from the model families GPT-2 \cite{radford2019language}, GPT-J \cite{gpt-j}, and Pythia \cite{biderman2023pythia}, implemented in Huggingface's Transformers \cite{wolf-etal-2020-transformers} for Python.
NER F1 scores are computed using Seqeval \cite{seqeval}.\\

\noindent
\textbf{Probing Classifiers.} For our probing classifiers, we use a multilayer perceptron (MLP) with a single hidden layer with $n_{\text{neurons}} \in \{32, 1024, 4096\}$ neurons and ReLU \cite{relu} as activation function.
We find that across all experiments the best results are obtained with $n_{\text{neurons}}=4096$.
We use cross-entropy loss and AdamW \cite{loshchilov_decoupled_2019} as optimizer, batch sizes $\in [1024, 4096]$ (learning rates $\in [5\mathrm{e}{-4},1\mathrm{e}{-4},5\mathrm{e}{-5}]$) and train using linear warmup (1 epoch) \cite{goyal_accurate_2017} followed by a linear learning rate decay.
We train tokenwise typing classifiers for $25$ epochs and span detection classifiers for $50$ epochs.\\

\noindent
\textbf{Data formatting.} We evaluate results at a tokenized level, meaning that we convert both the texts as well as the labels for CoNLL2003 and Ontonotes5 using the appropriate tokenizers for a given LM.
When training probing classifiers, we do not structure our batches according to source data samples, but instead at a ``per representation'' level:
In our implementation we begin by sampling internal LM representations for each token (or attention weight) in the NER dataset and cache the representations.
During the training of the probes, we sample from these representations, meaning that for tokenwise classification, a batch size of $n$ corresponds to $n$ hidden states, not $n$ training example texts from CoNLL2003 or Ontonotes5.\\

\noindent
\textbf{Few-Shot Learning.} 
We construct the few-shot task similar to the standard $n$-way $k$-shot setting with $n=4$ dictated by the amount of classes given in the dataset.
We evaluate each model for 200 episodes, the support set for each of which is sampled by retrieving $k$ data samples containing at least one mention of each entity type.
If a sample contains multiple entity mentions, we also count these towards $k$.
In order to use \methodname in this setting, we save the hidden states at a single layer\footnote{As our experiments show that deeper layers are more suitable for entity typing, we select a layer two thirds deep into the network.} and the attention weights between all tokens in all support data samples.
Instead of training probing classifiers, we then perform nearest neighbour classification based on the support representations.

\section{Detailed Label Propagation and Span Detection Results for all 7 LMs}
\label{sec:appendix-spandetection}
The results of the different label propagation strategies outlined in section \ref{sec:approach-pipeline} for all models are given in table \ref{tab:results-ablation-full-conll} for CoNLL2003 and table \ref{tab:results-ablation-full-ontonotes} for Ontonotes5.
The results of the different span detection strategies outlined in section \ref{sec:approach-spandetection} are given in table \ref{tab:results-spandetection}.
For adjacency classification, we see F1 scores of up to $98.43\%$ on CoNLL2003 and up to $93.23\%$ for Ontonotes5.
For the two span classification approaches, we find that predicting $f_{span}(A_{k, i}) = \hat{s}_{i,j}$ based on $j=k-1$ (``next'') outperforms the alternative, with up to $94.2\%$ for CoNLL2003 and $83.92\%$ for Ontonotes5.
Note that these results are obtained on individual data samples (individual representations paired with labels, as used during training) so that the evaluation and metrics calculation is not computed at the sequence level.
Therefore these results are not comparable with mention detection scores given in the other experiments, as those are computed using the full \methodname pipelines and at the sequence level.

\section{Extended Supervised Learning Benchmark Results}
\label{sec:appendix-full-results}
\begin{table*}[t]
\centering
\scalebox{0.85}{
\begin{tabular}{l|c|ccc|c|ccc}
\textbf{Model} & \multicolumn{4}{c|}{\textbf{CoNLL2003}} & \multicolumn{4}{c}{\textbf{Ontonotes5}} \\ \hline
 & $l$ & P & R & F1 & $l$ & P & R & F1 \\
\hline
GPT-2$_\text{small}$ & 8/12 & 86.66\% & 72.49\% & 78.94\% & 12/12 & 85.48\% & 62.02\% & 71.89\% \\
GPT-2$_\text{XL}$ & 36/48 & 89.01\% & 81.59\% & 85.14\% & 30/48 & 87.80\% & 72.24\% & 79.26\% \\
\hline
GPT-J$_\text{6B}$ & 12/28 & 89.54\% & 78.54\% & 83.68\% & 16/28 & 87.07\% & 68.54\% & 76.70\% \\
\hline
Pythia$_\text{410m}$ & 15/24 & 87.42\% & 76.63\% & 81.67\% & 14/24 & 86.08\% & 68.90\% & 76.54\% \\
Pythia$_\text{1.4b}$ & 18/24 & 88.54\% & 76.84\% & 82.27\% & 16/24 & 87.13\% & 68.33\% & 76.59\% \\
Pythia$_\text{2.8b}$ & 21/32 & 89.58\% & 80.21\% & 84.63\% & 17/32 & 87.66\% & 72.05\% & 79.09\% \\
Pythia$_\text{6.9b}$ & 22/32 & 89.37\% & 79.05\% & 83.90\% & 19/32 & 87.44\% & 71.80\% & 78.85\% \\

\end{tabular}
}
\caption{Full NER F1 scores for CoNLL2003 and Ontonotes5 using EMBER (spanwise label propagation) in the supervised learning setting. $l$ denotes the layer index at which the hidden states are probed for entity typing (chosen via hyperparameter optimization).}
\label{tab:results-full-appendix}
\end{table*}

See table \ref{tab:results-full-appendix} for precision, recall, and F1 scores for supervised learning benchmarks across all 7 LMs, as well as the layer index $l$ chosen for entity typing via hyperparameter optimization.

\section{Classification Examples}
\label{sec:appendix-examples}

Figures \ref{fig:example-output} and \ref{fig:example-output-hp} show examples of prompt annotated with \methodname on GPT-2$_\text{XL}$ and trained on Ontonotes5.
The prompt in figure \ref{fig:example-output} was chosen to show how predicted entity types change as more context information is incorporated into long spans (\textit{``The 57th New York Film Festival''}), which is previously unavailable to the model due to autoregressivity.
The prompt in figure \ref{fig:example-output-hp} was chosen to highlight an inherent limitation of \methodname due to autoregressivity which can not be fixed using the proposed methods.

\section{Model Architecture Details}
\label{sec:appendix-modelspecs}
\begin{table}[t]
\centering
\scalebox{0.8}{
\begin{tabular}{lcccc}
\textbf{Model} & \textbf{Hidden dim} & \textbf{\# att heads} & \textbf{\# layers} & \textbf{$|A|$} \\ \hline
\multicolumn{5}{c}{\textit{GPT-2}} \\ \hline
GPT-2$_\text{small}$ & 768 & 12 & 12 & 144 \\
GPT-2$_\text{XL}$ & 1600 & 25 & 48 & 1200 \\ \hline
\multicolumn{5}{c}{\textit{GPT-J}} \\ \hline
GPT-J$_\text{6B}$ & 4096 & 16 & 28 & 448 \\ \hline
\multicolumn{5}{c}{\textit{Pythia}} \\ \hline
Pythia$_\text{410m}$ & 1024 & 16 & 24 & 384 \\
Pythia$_\text{1.4b}$ & 2048 & 16 & 24 & 384 \\
Pythia$_\text{2.8b}$ & 2560 & 32 & 32 & 1024 \\
Pythia$_\text{6.9b}$ & 4096 & 32 & 32 & 1024 \\
\end{tabular}}
\caption{Relevant architecture parameters for models used in experiments.}
\label{tab:model-specs}
\end{table}

In table \ref{tab:model-specs}, we detail the relevant architecture parameters of the models used in our experiments.

\section{Entity Typing based on Last Token}
\label{sec:appendix-entitytyping}

\begin{table}
\centering
\scalebox{0.85}{
\begin{tabular}{l|cc}
\textbf{Model} & conll2003 & ontonotes5 \\ \hline

 GPT-2$_\text{small}$ & 93.99\% & 92.03\% \\
 GPT-2$_\text{XL}$ & 95.46\% & 92.73\% \\
 GPT-J$_\text{6B}$ & 96.25\% & 93.45\% \\
 Pythia$_\text{410m}$ & 94.98\% & 92.75\% \\
 Pythia$_\text{1.4b}$ & 95.59\% & 93.08\% \\
 Pythia$_\text{2.8b}$ & 95.93\% & 92.86\% \\
 Pythia$_\text{6.9b}$ & 96.38\% & 93.02\% \\

\end{tabular}
}
\caption{Entity typing F1 scores (based on last token of a span).}
\label{tab:results-entitytyping}
\end{table}

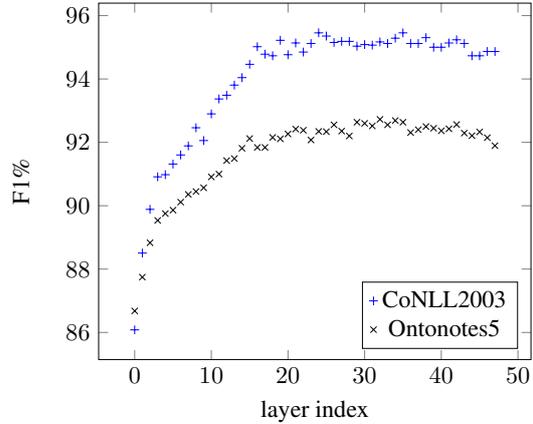
\begin{figure}[t]
\centering
\scalebox{0.83}{
    \begin{tikzpicture}
    \begin{axis}[legend pos=south east, ylabel=F1\%, xlabel=layer index, only marks, xshift=-5cm]
    \addplot[mark=+, color=blue] table [x=x, y=y, col sep=comma] {plot_layers_conll.csv};\addlegendentry{CoNLL2003}
    \addplot[mark=x, color=black] table [x=x, y=y, col sep=comma] {plot_layers_onto.csv};\addlegendentry{Ontonotes5}
    \end{axis}
    \end{tikzpicture}
    }
\caption{Entity typing F1 scores (validation set) for GPT-2$_\text{XL}$ with respect to the chosen layer.}
    \label{fig:layerwise}
\end{figure}

In table \ref{tab:results-entitytyping} we show results for entity typing where, given the correct spans, we use only the last token of each span to predict its type.
We measure F1 scores of up to $96.38\%$ and $93.45\%$, which we argue supports our choice of using a span's last token for label propagation.\\

Figure \ref{fig:layerwise} shows a plot of entity typing F1 scores measured using the hidden states at different layers of GPT-2$_\text{XL}$ as feature space.
We observe a clear trend showing that representations at earlier layers are less suitable for entity typing, which is in line with the findings of similar previous studies.
It is also the basis for choosing layers $2/3$ deep into the LM for the few-shot experiments.\\

\section{Few-Shot NER for GPT-2$_\text{small}$ and Pythia Models}
\label{sec:appendix-fewshot}
\begin{table}[t]
\centering
\scalebox{0.75}{
\begin{tabular}{l|ccccc}
\textbf{Model} & \textbf{1-shot} & \textbf{5-shot} & \textbf{10-shot} & \textbf{50-shot} & \textbf{100-shot} \\ \hline
GPT-2$_\text{small}$ & 23.67\% & 34.02\% & 37.24\% & 43.81\% & 48.11\% \\
Pythia$_\text{410m}$ & 22.52\% & 34.09\% & 38.79\% & 45.07\% & 46.45\% \\
Pythia$_\text{1.4b}$ & 23.23\% & 37.91\% & 42.44\% & 50.75\% & 53.72\% \\
Pythia$_\text{2.8b}$ & 23.52\% & 38.29\% & 41.72\% & 51.77\% & 54.66\% \\
Pythia$_\text{6.9b}$ & 14.11\% & 25.07\% & 32.90\% & 41.83\% & 47.93\% \\

\end{tabular}
}
\caption{Few-Shot F1 scores for NER on CoNLL2003. All scores are micro F1 scores.}
\label{tab:results-fewshot-pythia}
\end{table}

Table \ref{tab:results-fewshot-pythia} shows the few-shot learning results for the remaining models not shown in table \ref{tab:results-fewshot}.
We observe that Pythia$_\text{6.9b}$ appears to be an outlier exhibiting particularly low F1 scores. This suggests that there are other factors at play in this particular setting which can not be explained given the variables we measure.

\section{Generated NER Datasets}
\label{sec:appendix-generated-dataset}
We construct the datasets for section \ref{sec:simgenNER} as follows:\\

\noindent
\textbf{Evaluation Dataset.} 
We begin by randomly sampling 50 texts from CoNLL2003.
Next we use each text as a prompt for GPT-2$_{\text{XL}}$ to generate 100 tokens using greedy decoding with a repetition penalty of $1.2$.
We manually annotate the generated texts including their prompts (in order to ensure that any potential differences in annotation style do not interfere with the comparison of prompts vs. generated texts) according to the annotation guidelines\footnote{\url{https://www.cnts.ua.ac.be/conll2003/ner/annotation.txt}} used for CoNLL2003.
The resulting dataset contains 91 entity mentions in the prompts and 297 entity mentions in the generated text.\\

\noindent
\textbf{Synthetic Training Dataset.}
We generate texts for the training and validation splits of CoNLL2003 in the same way as for the evaluation dataset (excluding the 50 samples used for evaluation from the validation split).
Instead of manual annotation, we annotate the texts using the reference model\footnote{\url{https://huggingface.co/FacebookAI/xlm-roberta-large-finetuned-conll03-english}}.
We train the span detection probe only on the generated portion of each text, masking out the prompt during feature generation.
The remaining training procedure is identical to that used in the supervised learning setting (\ref{sec:supervised}).

\section{Entity Typing and Mention Detection Scores for Generated Text}
\label{sec:appendix-generation-isolated}
\begin{table}[t]
\centering
\scalebox{0.75}{
\begin{tabular}{lccc}
\textbf{Data} & \textbf{P} & \textbf{R} & \textbf{F1} \\ \hline
\multicolumn{4}{c}{\textsc{entity typing}} \\
\hline
original & $87.91\%$ & $87.91\%$ & $87.91\%$ \\
generated & $88.22\%$ & $88.22\%$ & $88.22\%$ \\
$\Delta$ & \textcolor{black}{$+0.30\%$} & \textcolor{black}{$+0.30\%$} & \textcolor{black}{$+0.30\%$} \\
\hline
\multicolumn{4}{c}{\textsc{md - \methodnamelower}} \\
\hline
original & $91.95\%$ & $87.91\%$ & $89.89\%$ \\
generated & $92.86\%$ & $70.03\%$ & $79.85\%$ \\
$\Delta$ & \textcolor{black}{$+0.90\%$} & \textcolor{red}{$-17.88\%$} & \textcolor{red}{$-10.04\%$} \\
\hline
\multicolumn{4}{c}{\textsc{md - \methodnamelower trained on generated}} \\
\hline
original & $89.87\%$ & $78.02\%$ & $83.53\%$ \\
generated & $92.83\%$ & $78.45\%$ & $85.04\%$ \\
$\Delta$ & \textcolor{black}{$+2.96\%$} & \textcolor{black}{$+0.43\%$} & \textcolor{black}{$+1.51\%$} \\
\hline
\multicolumn{4}{c}{\textsc{md - reference}} \\
\hline
original & $91.40\%$ & $93.41\%$ & $92.39\%$ \\
generated & $85.43\%$ & $86.87\%$ & $86.14\%$ \\
$\Delta$ & \textcolor{red}{$-5.97\%$} & \textcolor{red}{$-6.54\%$} & \textcolor{red}{$-6.25\%$} \\
\end{tabular}}
\caption{Isolated entity typing and mention detection scores measured on generated and non-generated data during experiments outlined in \ref{sec:simgenNER}.}
\label{tab:mention-detection-generated}
\end{table}

In table \ref{tab:mention-detection-generated} we show entity typing and mention detection scores for generated text in isolation. 
This data clearly shows that the drop in performance is due to mention detection recall suffering for \methodname trained on non-generated text.
Based on these results we retrain only the span detection probe on synthetically annotated generated text.

\section{Attention Windowing during Generation}
\label{sec:appendix-windowing}
\begin{table}[t]
\centering
\scalebox{0.75}{
\begin{tabular}{lccc}
\textbf{Data} & \textbf{P} & \textbf{R} & \textbf{F1} \\ \hline
\multicolumn{4}{c}{\textsc{\methodnamelower (gpt-2$_\text{XL}$) - windowed}} \\
\hline
original & $83.72\%$ & $79.12\%$ & $81.36\%$ \\
generated & $85.04\%$ & $67.00\%$ & $74.95\%$ \\
$\Delta$ & \textcolor{black}{$+1.32\%$} & \textcolor{red}{$-12.12\%$} & \textcolor{red}{$-6.40\%$} \\
\hline
$\Delta$$_\text{MD}$ & \textcolor{red}{$-1.57\%$} & \textcolor{red}{$-15.86\%$} & \textcolor{red}{$-9.79\%$} \\
\end{tabular}}
\caption{NER scores for windowed attention weights (window size $10$).}
\label{tab:generated-windowed}
\end{table}

During our experiments in simultaneous generation and extraction we hypothesized that another factor, rather than different attention behaviour on generated text, could have caused a model trained on non-generated text to perform poorly on generated text:
The generated texts are necessarily longer than the original texts (prompt + generation). 
Since mention detection is performed based only on the softmax normalized attention weights between two tokens, attention weights may be lower for longer contexts.
We, therefore, repeated the measurements obtained with the model trained on non-generated data, this time masking attention weights between tokens at a distance (with the exception of the attention weight directed at token 0 as this is often high regardless of distance) higher than $10$, and find that the drop in recall persists (albeit reduced).
The measured results are given in table \ref{tab:generated-windowed} and prompted us to reject this hypothesis.

\section{Results for WNUT2017 and BC5CDR}
\label{sec:appendix-further-datasets}
\begin{table}[h]
\centering
\scalebox{1}{
\begin{tabular}{l|cc}
\textbf{Model} & \multicolumn{1}{c}{\textbf{WNUT2017}} & \multicolumn{1}{c}{\textbf{BC5CDR}} \\ \hline
\hline
BINDER$_\text{distant}$  & - & 81.6\% \\
BINDER$_\text{supervised}$  & - & 91.9\% \\
CL-KL  & 60.45\% &  - \\
\hline
\methodname - GPT-2$_\text{XL}$  & 35.44\% & 75.07\% \\

(mention detection) & 39.80\% & 75.45\% \\

(entity typing)  & 61.08\% & 98.89\% \\

\end{tabular}
}
\caption{NER scores for WNUT2017 and BC5CDR using gpt2-xl. All scores are micro F1 scores. Results for BINDER are cited from \citet{zhang2023optimizing} and results for CL-KL are cited from \citet{wang-etal-2021-improving}.}
\label{tab:results-rebuttal}
\end{table}

In Table \ref{tab:results-rebuttal} we show the results measured using \methodname with GPT-2$_{\text{XL}}$ for the datasets WNUT2017 and BC5CDR.
As with CoNLL2003 and Ontonotes5, we find that mention detection is the major bottleneck for our approach.

\section{Entity Typing and Mention Detection Scores}
\label{sec:appendix-et-v-md}

\begin{table}[!phtb]
\centering
\scalebox{0.93}{
\begin{tabular}{l|cc}
\textbf{Dataset} & \multicolumn{1}{c}{\textbf{Entity Typing}} & \multicolumn{1}{c}{\textbf{Mention Detection}} \\
\hline
CoNLL2003 & 95.46\% & 94.20\% \\
Ontonotes5 & 92.73\% & 83.92\% \\
WNUT2017 & 61.08\% & 39.80\% \\
BC5CDR & 98.89\% & 75.45\% \\

\end{tabular}
}
\caption{Entity typing and mention detection scores for \methodname using gpt2-xl. All scores are micro F1 scores.}
\label{tab:results-et_v_md}
\end{table}

In Table \ref{tab:results-et_v_md} we show the entity typing and mention detection scores for \methodname with GPT-2$_{\text{XL}}$ for 4 datasets.

\section{Streaming Token Classification Implementation and Toolkit}
\label{sec:appendix-STOKE}

\begin{figure}[htbp]
  \centering
  \includegraphics[width=0.35\textwidth]{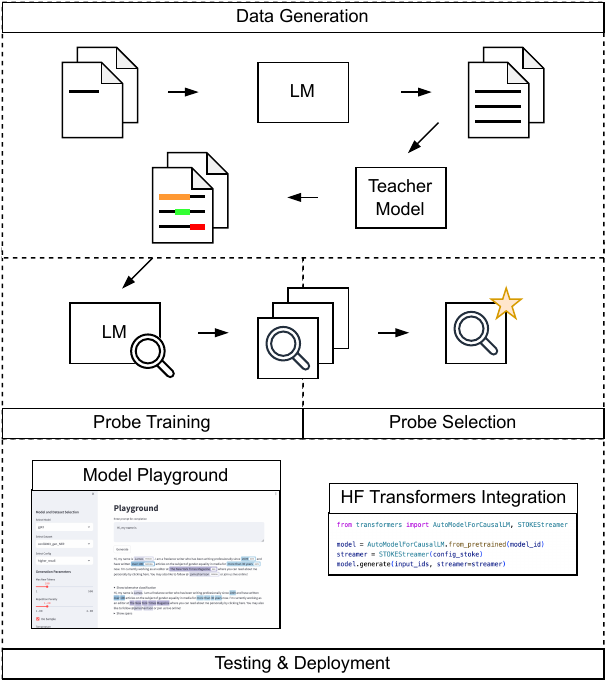}
  \caption{Overview of the workflow and tools implemented in the toolkit.}
  \label{fig:STOKE_overview}
\end{figure}

Viewed purely from an implementation perspective, any token classification task can be performed in the same way as \methodname (naturally, the constraint of autoregressivity will exclude some tasks as sensible candidates).
We therefore developed a toolkit for training, testing, and deploying custom streaming token classification models and workflows.

\subsection{Toolkit Design \& Implementation}

The toolkit, an overview of which is shown in figure \ref{fig:STOKE_overview}, includes: (1) a data generation pipeline, which follows a knowledge distillation approach for generating texts using language models and annotating them using teacher models, (2) a training and hyperparameter optimization pipeline, (3) code for the integration of trained classifiers into the Huggingface transformers ecosystem, and (4) a streamlit-based model playground for testing and debugging of classifiers.
We make all code, as well as a set of pre-trained classifiers available online at \url{https://github.com/nicpopovic/stoke}.

In this section we outline the different components of the toolkit (fig. \ref{fig:STOKE_overview}).
In order to avoid confusion, we clarify the model types involved in the workflow:
The language model (\textbf{LM}) is the decoder-only pre-trained language model, for which the user wants to incorporate streaming token classification. It remains unchanged throughout the entire process.
The \textbf{teacher model} is an auxiliary model which has been trained to perform the type of token classification the user wants to integrate into the LM.
The \textbf{probing classifiers} are multilayer perceptrons with a single hidden layer each and are trained to perform the token classification task on the internal representations of the LM.

\subsubsection{Data Generation}

In particular for span detection, training probes on text generated by the LM provides better results than using non-generated text for training.
Thus, the first step of the data pipeline is the generation of texts using the LM and a set of prompts.
We note that while arbitrary prompts can be used in this step, the generated text is a product of the prompts. Choosing prompts as close as possibly to the target domain will, therefore, likely yield better results than random prompts.
In our experiments, we therefore use datasets which have been constructed for a given task as our prompts, for example we use the texts provided in CoNLL2003 \cite{tjongkimsang2003conll} as prompts for the NER task.
Having generated a text corpus, the next step is to annotate it with respect to the target task using a teacher model.
For the initial set of tasks, we use the FLAIR framework \cite{akbik2019flair}, which includes pretrained models for NER, POS-tagging, chunking, and verb disambiguation.
Further tasks can be easily integrated using the pipelines supplied in Huggingface's Transformers \cite{wolf-etal-2020-transformers}.
Finally, the produced dataset is split into subsets, for training, validation, and testing.
The above workflow has been implemented to be run from a single command for language models available in the Transformers library.

\subsubsection{Model Training}

The model training pipeline, also run with a single command, iterates over the training dataset generated in the previous step, feeds each training example into the LM in a forward pass and trains probing classifiers to predict the labels based on the LMs internal representations.
Since the probing classifiers are typically small (up to 4096 hidden units in our experiments), the pipeline is designed to train multiple probing classifiers simultaneously with each forward pass of the LM.
We therefore implement a simple grid-search based hyperparameter optimization strategy.
We note that the implementation of our toolkit with respect to batching during training differs from the procedure used in the experiments conducted for this paper due to efficiency reasons.
It is currently not known to what extent this effects the results.
For more details, we refer to the implementation details in appendix \ref{sec:appendix-implementation-details} and the code we provide for both our experiments and the toolkit.

\subsubsection{Model Evaluation \& Selection}
Depending on the hyperparameter ranges selected during model training, hundreds of probing classifiers may have been trained.
The evaluation pipeline selects two classifiers using the following strategy:
For the token classification probe we select the one with the highest F1 score on the development set (measured during training), as typically more token classifiers ($f_\text{token}$) have been trained than span classifiers.
Then, using a dataset held out for testing, the span classifier ($f_\text{span}$) which results in the highest F1 score when used in conjunction with the selected token classifier is chosen for the final configuration.
Again, this evaluation step is called by the user with a single command and outputs a configuration file which can then be used in testing and deployment.

\subsubsection{Testing \& Deployment}
For testing and deployment, the toolkit includes the code necessary to easily integrate the trained models into existing applications based on the Transformers \cite{wolf-etal-2020-transformers} Python library.
At the time of writing, the Transformers library only passes generated tokens to streamers\footnote{\url{https://github.com/huggingface/transformers/blob/f6261d7d81edd036fc53bfede65fe91f01a661aa/src/transformers/generation/utils.py\#L2458}}, while our method requires hidden states and attention weights.
We provide a fork\footnote{\url{https://github.com/nicpopovic/transformers}} of the library with the minimal necessary changes and documentation for how to apply them to other versions of the library.

\begin{figure*}
  \centering
  \includegraphics[width=\textwidth]{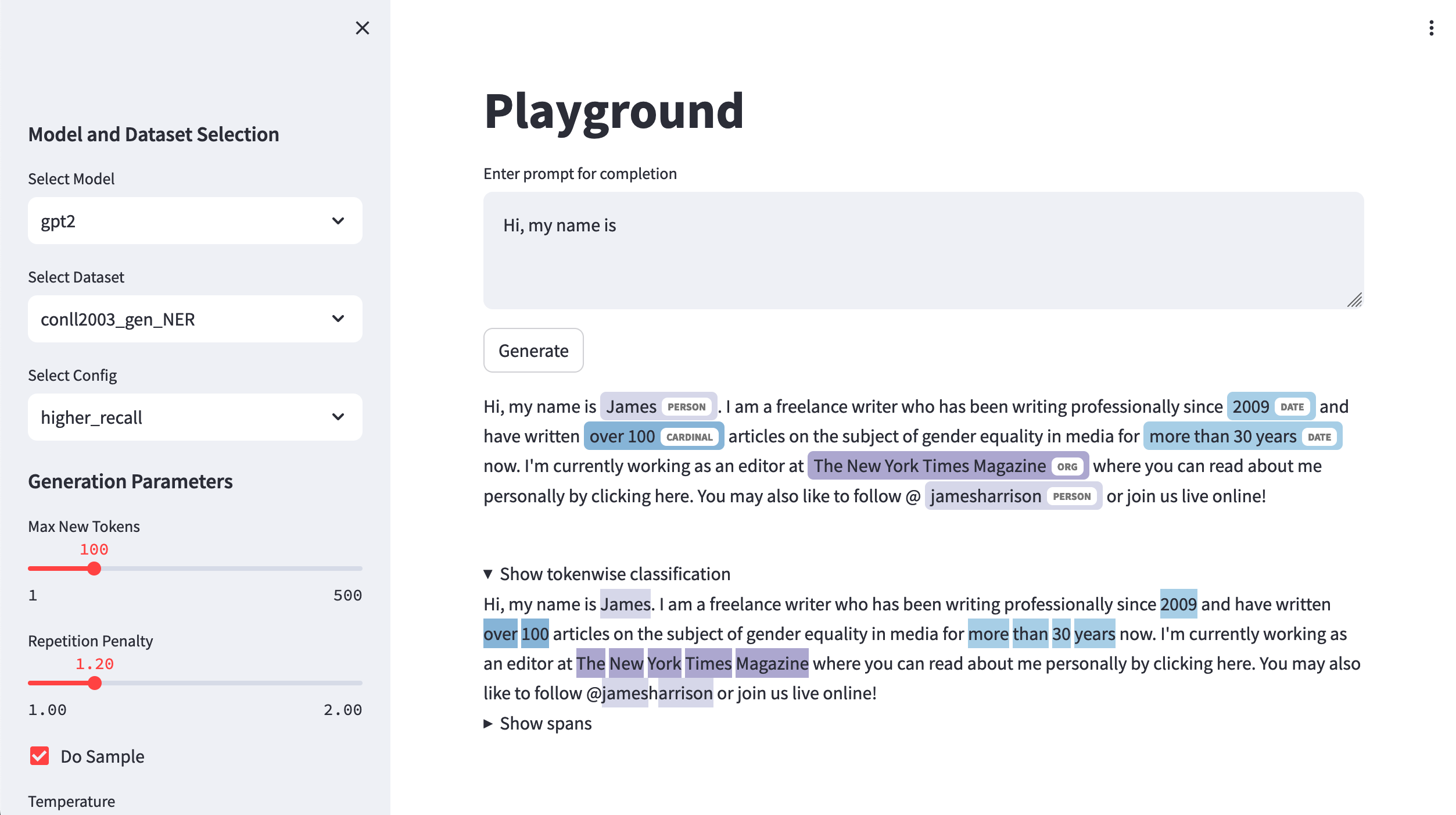}
  \caption{Screenshot of the model playground.}
  \label{fig:playground}
\end{figure*}

For the purpose of qualitative testing of the trained classifiers, we provide a model playground in the form of a web application implemented in Python using Streamlit (\url{https://github.com/streamlit/streamlit}).
A screenshot of this model playground is shown in figure \ref{fig:playground}.
The interface lets the user choose from the different models, tasks, and combinations of probing classifiers produced in the pipeline.
It includes a sidebar for choosing various generation and classification parameters, as well as the main prompt and output views.
After choosing the desired parameters, a user enters a prompt and clicks the ``generate'' button.
Text is generated using the selected model and parameters and is annotated using the chosen classifier settings.
The classified token sequence is streamed to the front-end, where the user can view both the final classification, as well as the tokenwise type classification.

\subsection{Illustration of Streaming Token Classification}

Finally, in order to illustrate the process of streaming token classification, we include an example of outputs generated by the individual components of a streaming token classification pipeline at each generation step in table \ref{tab:example_outputs}.

\section{Applicability to Newer Models}
\label{sec:appendix-llama}
In order to examine how well \methodname works when applied to more current LMs, we evaluated two more recent models, specifically Llama3.2$_\text{1b}$ and Llama3.2$_\text{3b}$ \cite{dubey2024llama3herdmodels}.
The results, shown in tables \ref{tab:results-llama-conll} and \ref{tab:results-llama-ontonotes}, indicate that performance is on par with the results seen for GPT-2$_\text{XL}$.
Both models have fewer attention heads compared to GPT-2$_\text{XL}$ ($512$ for Llama3.2$_\text{1b}$ and 672 for Llama3.2$_\text{3b}$), indicating that other factors than the number of attention heads can also benefit NER capabilities.

\begin{table}[h]
\centering
\scalebox{1}{
\begin{tabular}{l|ccc}
\textbf{Model} & \multicolumn{3}{c}{\textbf{CoNLL2003}}\\ \hline
 & P & R & F1 \\

\hline
GPT-2$_\text{XL}$  & 89.01\% & 81.59\% & 85.14\% \\
Llama3.2$_\text{1b}$  & 89.39\% & 82.32\% & 85.71\% \\
Llama3.2$_\text{3b}$  & 90.21\% & 82.23\% & 86.04\% \\

\end{tabular}
}
\caption{Evaluation of \methodname applied to Llama3.2 for CoNLL2003. All scores are micro F1 scores.}
\label{tab:results-llama-conll}
\end{table}

\begin{table}[h]
\centering
\scalebox{1}{
\begin{tabular}{l|ccc}
\textbf{Model} & \multicolumn{3}{c}{\textbf{Ontonotes5}}\\ \hline
 & P & R & F1 \\

\hline
GPT-2$_\text{XL}$ & 87.80\% & 72.24\% & 79.26\% \\
Llama3.2$_\text{1b}$ & 87.27\% & 71.46\% & 78.58\% \\
Llama3.2$_\text{3b}$ & 87.68\% & 71.96\% & 79.05\% \\

\end{tabular}
}
\caption{Evaluation of \methodname applied to Llama3.2 for Ontonotes5. All scores are micro F1 scores.}
\label{tab:results-llama-ontonotes}
\end{table}

\begin{table}[!htb]
\centering
\scalebox{0.85}{
\begin{tabular}{l|cc|c}
 & \multicolumn{2}{c|}{span detection} &  \\
\textbf{Model} & last & next & adjacency \\ \hline

\hline
\multicolumn{4}{c}{\textsc{conll2003}} \\
\hline
GPT-2$_\text{small}$ & 85.49\% & \textbf{89.65\%} & 96.62\%\\
GPT-2$_\text{XL}$ & 91.07\% & \textbf{94.20\%} & 97.64\%\\
GPT-J$_\text{6B}$ & 90.11\% & \textbf{91.59\%} & 98.00\%\\
Pythia$_\text{410m}$ & 88.50\% & \textbf{91.04\%} & 97.69\%\\
Pythia$_\text{1.4b}$ & 88.89\% & \textbf{91.37\%} & 97.84\%\\
Pythia$_\text{2.8b}$ & 90.83\% & \textbf{93.02\%} & 98.25\%\\
Pythia$_\text{6.9b}$ & 90.94\% & \textbf{92.99\%} & 98.43\%\\

\hline
\multicolumn{4}{c}{\textsc{ontonotes5}} \\
\hline
GPT-2$_\text{small}$ & 75.14\% & \textbf{76.56\%} & 89.11\%\\
GPT-2$_\text{XL}$ & 81.46\% & \textbf{83.92\%} & 91.56\%\\
GPT-J$_\text{6B}$ & 79.28\% & \textbf{81.27\%} & 91.90\%\\
Pythia$_\text{410m}$ & 77.07\% & \textbf{80.71\%} & 92.12\%\\
Pythia$_\text{1.4b}$ & 78.26\% & \textbf{80.73\%} & 92.48\%\\
Pythia$_\text{2.8b}$ & 79.63\% & \textbf{82.93\%} & 93.21\%\\
Pythia$_\text{6.9b}$ & 79.99\% & \textbf{83.48\%} & 93.23\%\\

\end{tabular}
}
\caption{\textit{Span detection:} Micro F1 scores (validation set) for mention detection classifiers trained on attention weights between either last or next token and the first token of a span. \textit{Adjacency:} Micro F1 scores (validation set) for classifiers using attention weights to classify whether two adjacent tokens belong to the same entity.}
\label{tab:results-spandetection}
\end{table}

\begin{table*}
  \centering
  \caption{Example outputs at each step of the streaming token classification. The language model outputs the most likely next token, the token type classifier outputs a type prediction for the most recent token of the context ($t_i$, underlined), and the span detection classifier outputs all detected spans between the last token of the context and any previous token. Finally, the tokenwise predictions and the detected spans are aggregated into final predictions.}
  \label{tab:example_outputs}
  \scalebox{0.5}{
  \begin{tabular}{c|r|l|c|c|l} %
    \toprule
     & \multicolumn{1}{c}{\textbf{Input}} & \multicolumn{4}{|c}{\textbf{Output}} \\
    \hline
    $i$ & \textbf{Context} & \textbf{Next token} & $f_\text{token}$ & $f_\text{span}$ & \textbf{Predictions} \\
    \hline
    0 &\underline{`Paul'} &  ` Atreides'  & `Paul' -> PERSON   & -   & -   \\
    1 &`Paul', \underline{` Atreides'}&  ` is'  & ` Atreides' -> PERSON   &  (0,0) -> 0.7  & (PERSON, (0,0), `Paul')   \\
    2 &`Paul', ` Atreides', \underline{` is'}&  ` the'  & ` is' -> O   & (0,1) -> 0.9   & (PERSON, (0,1), `Paul Atreides')   \\
    3 &`Paul', ` Atreides', ` is', \underline{` the'} &  ` protagonist'  & ` the' -> O & - & (PERSON, (0,1), `Paul Atreides')   \\
    4 &`Paul', ` Atreides', ` is', ` the', \underline{` protagonist'} &  ` of'  & ` protagonist' -> O   & -   & (PERSON, (0,1), `Paul Atreides')   \\
    5 &`Paul', ` Atreides', ` is', ` the', ` protagonist', \underline{` of'} &  ` "'  & ` of' -> O   & (4,5) -> 0.6   & (PERSON, (0,1), `Paul Atreides')   \\
    6 &`Paul', ` Atreides', ` is', ` the', ` protagonist', ` of', \underline{` "'} &  ` Dune'  & ` "' -> O   & -   & (PERSON, (0,1), `Paul Atreides')   \\
    7 &`Paul', ` Atreides', ` is', ` the', ` protagonist', ` of', ` "', \underline{`Dune'} &  ` "'  & ` Dune' -> WORK\_OF\_ART   & -   & (PERSON, (0,1), `Paul Atreides')   \\
    8 &`Paul', ` Atreides', ` is', ` the', ` protagonist', ` of', ` "', `Dune', \underline{`"'}&  ` and'  & ` "' -> O   & (7,7) -> 0.8   & (PERSON, (0,1), `Paul Atreides'), (WORK\_OF\_ART, (7,7), `Dune')   \\
    \bottomrule
  \end{tabular}
}
\end{table*}

\begin{table}[h]
\centering
\scalebox{1}{
\begin{tabular}{lccc}
\textbf{Model} & P & R & F1 \\ \hline

\hline
\multicolumn{4}{c}{\textsc{conll2003}} \\
\hline

\multicolumn{4}{c}{Tokenwise typing }\\
\hline
 GPT-2$_\text{small}$ & 62.50\% & 77.90\% & 69.36\% \\
 GPT-2$_\text{XL}$ & 64.55\% & 79.10\% & 71.09\% \\
 GPT-J$_\text{6B}$ & 65.88\% & 80.24\% & 72.36\% \\
 Pythia$_\text{410m}$ & 56.90\% & 74.87\% & 64.66\% \\
 Pythia$_\text{1.4b}$ & 63.12\% & 78.04\% & 69.79\% \\
 Pythia$_\text{2.8b}$ & 63.26\% & 78.14\% & 69.91\% \\
 Pythia$_\text{6.9b}$ & 64.46\% & 79.13\% & 71.05\% \\
\hline

\hline
\multicolumn{4}{c}{Adj. propagation }\\
\hline
 GPT-2$_\text{small}$ & 82.25\% & 87.68\% & 84.88\% \\
 GPT-2$_\text{XL}$ & 84.40\% & 90.47\% & 87.33\% \\
 GPT-J$_\text{6B}$ & 86.70\% & 92.02\% & 89.28\% \\
 Pythia$_\text{410m}$ & 82.86\% & 88.84\% & 85.75\% \\
 Pythia$_\text{1.4b}$ & 84.24\% & 90.39\% & 87.21\% \\
 Pythia$_\text{2.8b}$ & 86.54\% & 91.84\% & 89.11\% \\
 Pythia$_\text{6.9b}$ & 87.50\% & 92.58\% & 89.97\% \\
\hline

\hline
\multicolumn{4}{c}{Spanwise typing }\\
\hline
 GPT-2$_\text{small}$ & 85.07\% & 88.34\% & 86.67\% \\
 GPT-2$_\text{XL}$ & 89.32\% & 91.75\% & 90.52\% \\
 GPT-J$_\text{6B}$ & 88.82\% & 91.60\% & 90.19\% \\
 Pythia$_\text{410m}$ & 84.44\% & 87.04\% & 85.72\% \\
 Pythia$_\text{1.4b}$ & 87.37\% & 90.22\% & 88.77\% \\
 Pythia$_\text{2.8b}$ & 88.37\% & 90.17\% & 89.26\% \\
 Pythia$_\text{6.9b}$ & 89.13\% & 91.35\% & 90.23\% \\
\hline

\hline
\multicolumn{4}{c}{Span propagation }\\
\hline
 GPT-2$_\text{small}$ & 92.41\% & 79.49\% & 85.46\% \\
 GPT-2$_\text{XL}$ & 94.08\% & 87.13\% & 90.47\% \\
 GPT-J$_\text{6B}$ & 94.49\% & 83.76\% & 88.80\% \\
 Pythia$_\text{410m}$ & 92.72\% & 81.61\% & 86.81\% \\
 Pythia$_\text{1.4b}$ & 93.84\% & 82.78\% & 87.96\% \\
 Pythia$_\text{2.8b}$ & 94.30\% & 85.41\% & 89.63\% \\
 Pythia$_\text{6.9b}$ & 94.90\% & 86.05\% & 90.26\% \\
\hline

\end{tabular}
}
\caption{NER scores using hidden states and attention weights in different ways. All scores are micro F1 scores measured on the validation set of CoNLL2003.}
\label{tab:results-ablation-full-conll}
\end{table}

\begin{table}[h]
\centering
\scalebox{1}{
\begin{tabular}{lccc}
\textbf{Model} & P & R & F1 \\ \hline

\hline
\multicolumn{4}{c}{\textsc{ontonotes5}} \\
\hline

\hline
\multicolumn{4}{c}{Tokenwise typing }\\
\hline
 GPT-2$_\text{small}$ & 55.69\% & 69.78\% & 61.94\% \\
 GPT-2$_\text{XL}$ & 58.56\% & 71.55\% & 64.41\% \\
 GPT-J$_\text{6B}$ & 59.52\% & 72.54\% & 65.39\% \\
 Pythia$_\text{410m}$ & 54.65\% & 67.96\% & 60.58\% \\
 Pythia$_\text{1.4b}$ & 55.98\% & 69.84\% & 62.15\% \\
 Pythia$_\text{2.8b}$ & 55.93\% & 69.92\% & 62.15\% \\
 Pythia$_\text{6.9b}$ & 57.68\% & 71.12\% & 63.70\% \\
\hline

\hline
\multicolumn{4}{c}{Adj. propagation }\\
\hline
 GPT-2$_\text{small}$ & 64.39\% & 72.92\% & 68.39\% \\
 GPT-2$_\text{XL}$ & 68.25\% & 76.11\% & 71.96\% \\
 GPT-J$_\text{6B}$ & 69.64\% & 77.85\% & 73.52\% \\
 Pythia$_\text{410m}$ & 68.74\% & 76.78\% & 72.54\% \\
 Pythia$_\text{1.4b}$ & 70.21\% & 77.42\% & 73.64\% \\
 Pythia$_\text{2.8b}$ & 70.79\% & 78.10\% & 74.27\% \\
 Pythia$_\text{6.9b}$ & 70.27\% & 77.84\% & 73.86\% \\
\hline

\hline
\multicolumn{4}{c}{Spanwise typing }\\
\hline
 GPT-2$_\text{small}$ & 71.57\% & 76.07\% & 73.75\% \\
 GPT-2$_\text{XL}$ & 76.79\% & 76.26\% & 76.52\% \\
 GPT-J$_\text{6B}$ & 75.22\% & 75.28\% & 75.25\% \\
 Pythia$_\text{410m}$ & 73.11\% & 73.64\% & 73.37\% \\
 Pythia$_\text{1.4b}$ & 74.00\% & 73.69\% & 73.84\% \\
 Pythia$_\text{2.8b}$ & 74.79\% & 74.54\% & 74.67\% \\
 Pythia$_\text{6.9b}$ & 75.27\% & 78.84\% & 77.01\% \\
\hline

\hline
\multicolumn{4}{c}{Span propagation }\\
\hline
 GPT-2$_\text{small}$ & 85.21\% & 62.31\% & 71.98\% \\
 GPT-2$_\text{XL}$ & 87.26\% & 72.77\% & 79.36\% \\
 GPT-J$_\text{6B}$ & 86.82\% & 69.88\% & 77.43\% \\
 Pythia$_\text{410m}$ & 85.29\% & 68.60\% & 76.04\% \\
 Pythia$_\text{1.4b}$ & 85.76\% & 68.46\% & 76.14\% \\
 Pythia$_\text{2.8b}$ & 86.67\% & 71.22\% & 78.19\% \\
 Pythia$_\text{6.9b}$ & 86.81\% & 72.14\% & 78.80\% \\
\hline

\end{tabular}
}
\caption{NER scores using hidden states and attention weights in different ways. All scores are micro F1 scores measured on the validation set Ontonotes5.}
\label{tab:results-ablation-full-ontonotes}
\end{table}

\end{document}